\documentclass[11pt]{article}

\usepackage[final]{acl}

\usepackage{times}
\usepackage{latexsym}
\usepackage[T1]{fontenc}
\usepackage[utf8]{inputenc}
\usepackage{microtype}
\usepackage{inconsolata}
\usepackage{graphicx}
\usepackage{booktabs}
\usepackage{adjustbox}
\usepackage{multirow}
\usepackage{amsmath}
\usepackage{xcolor}
\usepackage{colortbl}
\usepackage{pifont}
\usepackage{caption}
\usepackage{subcaption}
\usepackage{tikz}
\usetikzlibrary{arrows.meta, positioning, calc, fit}
\usepackage[most]{tcolorbox}
\usepackage{enumitem}
\usepackage{stfloats}
\newtcolorbox{promptbox}[2][]{
  breakable,
  colback=gray!8,
  colframe=gray!40,
  boxrule=0.4pt,
  arc=1.5pt,
  left=4pt, right=4pt, top=3pt, bottom=3pt,
  fontupper=\small,
  title={#2},
  coltitle=black,
  colbacktitle=gray!25,
  fonttitle=\small\bfseries,
  #1
}

\newcommand{\smark}{\textsuperscript{*}}

\makeatletter
\newcommand\runinhead{\@startsection{paragraph}{4}{\z@}%
  {1.5ex \@plus 0.5ex \@minus .2ex}%
  {-0.4em}%
  {\normalsize\bfseries}}
\makeatother

\definecolor{win75}{HTML}{A5D6A7}
\definecolor{win65}{HTML}{C8E6C9}
\definecolor{win55}{HTML}{E8F5E9}
\definecolor{win45}{HTML}{FFF8E1}
\definecolor{win35}{HTML}{FFCCBC}

\definecolor{dpos8}{HTML}{A5D6A7}
\definecolor{dpos4}{HTML}{C8E6C9}
\definecolor{dpos0}{HTML}{E8F5E9}
\definecolor{dneg0}{HTML}{FFF8E1}
\definecolor{dneg4}{HTML}{FFCCBC}
\definecolor{dneg8}{HTML}{EF9A9A}

\definecolor{mygreen}{HTML}{2E7D32}
\definecolor{mygreenbg}{HTML}{E8F5E9}
\definecolor{mydarkgreen}{HTML}{1B5E20}
\definecolor{mydarkgreenbg}{HTML}{C8E6C9}
\definecolor{myorange}{HTML}{E87400}
\definecolor{myorangebg}{HTML}{FEF3DE}
\definecolor{mygray}{HTML}{546E7A}
\definecolor{mygraybg}{HTML}{CFD8DC}

\definecolor{cmetafill}{HTML}{D5D5D5}
\definecolor{cmetastroke}{HTML}{757575}
\definecolor{csrcfill}{HTML}{C5D2EA}
\definecolor{csrcstroke}{HTML}{4D6B9F}
\definecolor{cinstrfill}{HTML}{F7D8B0}
\definecolor{cinstrstroke}{HTML}{B97A3C}
\definecolor{ctransfill}{HTML}{C9E5BF}
\definecolor{ctransstroke}{HTML}{5C8A4D}
\definecolor{cllmstroke}{HTML}{3F3F3F}
\newcommand{\diffbox}[4]{%
    \scriptsize
    \tikz[baseline=(char.base)]\node[
        rounded corners=2pt, fill=#1, text=#2,
        inner sep=1.5pt, font=\bfseries,
    ] (char) {#3 #4};%
}
\newcommand{\goodup}[1]{\diffbox{mygreenbg}{mygreen}{$\uparrow$}{#1}}
\newcommand{\verygoodup}[1]{\diffbox{mydarkgreenbg}{mydarkgreen}{$\uparrow$}{#1}}
\newcommand{\gooddown}[1]{\diffbox{mygreenbg}{mygreen}{$\downarrow$}{#1}}

\title{Beyond ``To whom it may concern'': Tailoring Machine Translation to Audience and Intent}

\author{
  Raphael Merx \quad
  Ekaterina Vylomova \quad
  Trevor Cohn \\
  The University of Melbourne \\
  \texttt{rmerx@student.unimelb.edu.au} \\}

\begin{document}
\maketitle

\begin{abstract}

Translation quality depends on purpose: the same source text demands different translations depending on audience, tone, and communicative intent. Yet MT models and metrics treat translation as a fixed mapping from source to target.
LLMs enable users to explicitly specify purpose alongside source text, yet this capability has not been evaluated at scale.
We introduce a systematic evaluation of purpose-driven MT across 50~languages, 5 model sizes and 8~text domains. We find that
(1)~explicit instructions substantially improve translation adaptedness, with larger gains on informal domains (conversation, social media), for larger model sizes and for higher-resource languages;
(2)~instructions outperform semantically-matched few-shot examples and paragraph-level context;
(3)~traditional MT metrics fail to capture adaptation quality, often penalizing adapted translations;
(4)~when curated instructions are unavailable, models can self-generate them from surrounding document context, closing up to 80\% of the adaptedness gap to curated instructions.
Our results establish that purpose-adapted MT is a viable and measurable capability of LLMs, while highlighting the need for purpose-aware metrics.
\end{abstract}

\section{Introduction}
\label{sec:intro}

Translation theory has long argued that quality is relative to purpose. Skopos theory \cite{nord-1994-importance} holds that a translation's adequacy should be measured against its intended function, not against a universal standard of correctness. For example, the English phrase ``Excuse me'' can be translated in myriad different ways depending on intent and audience. Yet traditional MT systems, and the metrics that evaluate them, treat translation as a function from source to target, ignoring purpose.

This reduction is a methodological choice. Decades of parallel-corpus training and single-reference evaluation (optimizing for scores against one ``correct'' reference translation) systematically excluded pragmatic factors like target audience and communicative intent \cite{ma-etal-2025-pragmatics}.

\citet{carpuat-etal-2025-interdisciplinary} call for a paradigm shift: a translation should be faithful to its intended purpose, not merely to the source.
LLMs offer a potential solution, as they let users specify the intended context (audience, formality level, domain) alongside the source text.
But does this actually work? For which languages, model sizes, and domains? And what is the most effective way to convey purpose: explicit instructions, in-context examples, or surrounding document context?

Prior work on context-aware MT is limited in scope, examining either implicit context through document-level translation \cite{wang-etal-2023-document-level}, or 
explicit translation specifications on a single language pair and a single domain \cite{yamada-2023-optimizing,he-2024-prompting,sharkas_exploring_2025}. NLP research on contextual adaptation through LLMs focuses on narrow phenomena like formality or pronoun consistency \cite{jiang-etal-2023-discourse,choudhary-etal-2025-exploring}. The broader question of whether LLMs can systematically leverage user-provided specifications to produce purpose-adapted translations remains open.

We address the following research questions:
\begin{enumerate}
    \item[\textbf{RQ1}] To what extent can LLMs adapt translations to instructions, and how does this vary by model size, language resource level, and domain?
    \item[\textbf{RQ2}] How does instruction-based adaptation compare to few-shot in-context examples, and to document context?
    \item[\textbf{RQ3}] When curated instructions are unavailable, can models generate effective ones from surrounding document context?
\end{enumerate}

We evaluate on two translation benchmarks, across 50+~languages, 5~model sizes (Gemma-3-4B/12B/27B, Gemma-4-31B, Qwen3.5-27B), and 8~domains. We find that:

\begin{enumerate}
    \item Instructions improve translation adaptedness, with gains scaling with model size (\S\ref{sec:model-size}), and particularly high for informal domains (\S\ref{sec:domain});
    \item Instructions outperform semantically-matched few-shot examples and paragraph-level translation (\S\ref{sec:fewshot});
    \item Models can self-generate effective instructions from surrounding document context, closing up to 80\% of the adaptedness gap to curated gold instructions (\S\ref{sec:self-instr}).
\end{enumerate}

Together, these findings show that LLMs can operationalize purpose-aware translation, and offer concrete mechanisms for adapting outputs to user intent, a step toward a more user-centered paradigm in MT research \footnote{Code: \href{https://github.com/raphaelmerx/purpose-mt/}{github.com/raphaelmerx/purpose-mt}}.

\begin{table}[t]
\centering
\small
\begin{tabular}{p{0.93\columnwidth}}
\toprule
\textbf{French} \textit{(Metadata: conversation; informal register; chat with a friend (both female) in a messenger; sarcasm)} \\
\textbf{Source:} ``I am glad that you agree'' \\
\textbf{Instruction:} Translate as a casual, sarcastic message between female friends in a messenger chat. \\
\textbf{W/o:} \textit{Je suis content(e) que vous soyez d'accord.} \\
{\small\hspace{1.5em}$\rightarrow$ formal vous, relatively flat sentence that feels stiff} \\
\textbf{W/:} \textit{Ah, super que tu sois d'accord, finalement.} \\
{\small\hspace{1.5em}$\rightarrow$ casual tu, more natural and lighter tone} \\
\midrule
\textbf{Javanese} \textit{(Metadata: narration; informal register; Inspired by the fairy tale Timun Mas)} \\
\textbf{Source:} ``Once upon a time, there lived a husband and wife in a village.'' \\
\textbf{Instruction:} Translate as a traditional fairy tale narration. Use simple, evocative language suitable for children. \\
\textbf{W/o:} \textit{Dukuné, ana lanang lan wadon sing urip ing desa.} \\
{\small\hspace{1.5em}$\rightarrow$ matter-of-fact phrasing with no storytelling flavor} \\
\textbf{W/:} \textit{Ing jaman biyen, ana sawijining pasangan bojo sing urip ing desa.} \\
{\small\hspace{1.5em}$\rightarrow$ fairy-tale opener, evocative and child-friendly} \\
\bottomrule
\end{tabular}
\caption{Example translations from the 27B model showing register adaptation through instructions. W/o~= without instruction; W/~= with instruction.}
\label{tab:examples}
\end{table}

\begin{figure}[t]
\centering
\includegraphics[width=\columnwidth]{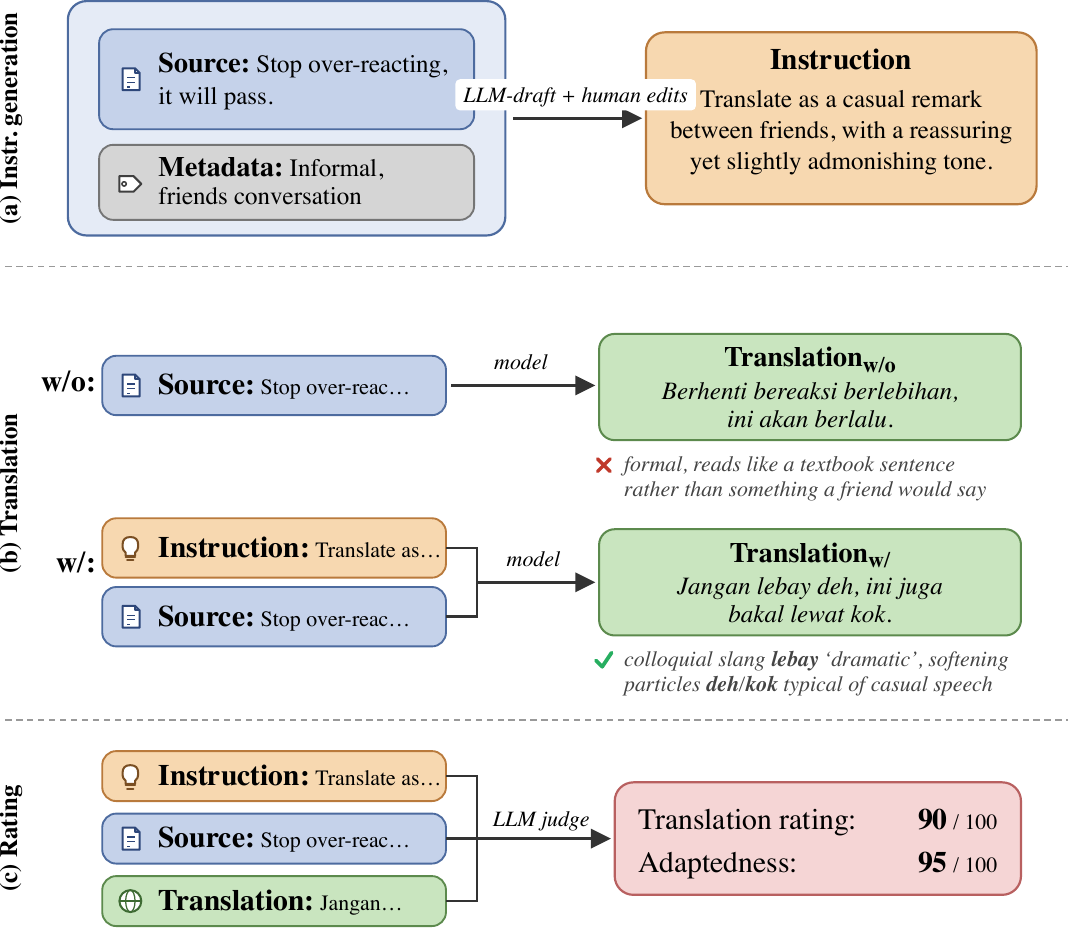}
\caption{Full pipeline with Indonesian example: (a)~instruction generation from source metadata; (b)~translation (without/with instructions); (c)~translation rating with an LLM-judge, which always sees the instructions. For additional examples, see Table~\ref{tab:examples}; for the full prompts used at each stage, see Appendix~\ref{app:prompts}.}
\label{fig:pipeline}
\end{figure}

\section{Related Work}
\label{sec:related}

\runinhead{Controllable MT}
focuses on steering MT output depending on pre-defined attributes: formality \cite{sennrich-etal-2016-controlling, niu-etal-2017-study, nadejde-etal-2022-cocoa}, gender (\citealt{vanmassenhove-etal-2018-getting, currey-etal-2022-mt}), text complexity \cite{agrawal-carpuat-2019-controlling}, and terminology \cite{semenov-etal-2023-findings}. These tasks pair each source with a category label (formal/informal, masculine/feminine, simple/complex).
Free-text instructions differ in two ways.
First, they let the model decide which category applies: if I'm addressing a stranger who is my age, should I use formal or informal language?
Second, they allow specifying intent for sense disambiguation (the ``Excuse me'' example in the Introduction).
Appendix~\ref{app:controlled-mt} evaluates on two controllable MT benchmarks (CoCoA-MT and MT-GenEval), and finds that implicit context (a free-text description of the situation) recovers 96\% of the accuracy that an explicit register tag provides. It also finds that models internalise language-specific conventions, such as addressing strangers online with an informal pronoun in German, but with a formal verb morphology in Japanese.

\runinhead{Specification-guided MT}
explores incorporating translation specifications (such as purpose, audience, or translation briefs) into LLM prompts, finding that specification-prompted translations receive higher human ratings \cite{yamada-2023-optimizing, he-2024-prompting, sharkas_exploring_2025, kayano-sugawara-2025-specification}. However, these evaluations are limited to single language pairs, small test sets (typically 3-5 passages), and often qualitative in nature.
For example, \citet{he-2024-prompting}, working with English-to-Chinese translation of a science article through GPT-4, find that a brief explanation of the translation purpose and context helps deliver a more appropriate translation.
Our work scales specification-guided MT evaluation to 50+~languages, 8~domains, 5~model sizes.

\paragraph{Contextual adaptation in LLM-based MT.}
Recent work investigates whether surrounding document context helps LLMs make discourse-level decisions such as formality consistency, pronoun agreement, and lexical cohesion \cite{choudhary-etal-2025-exploring}. A tension exists between in-context examples and explicit specifications, where LLMs can ignore explicit instructions when given few-shot examples \cite{zhu-etal-2024-multilingual}. More broadly, the question of whether LLMs translate better from explicit instructions, or implicit patterns found in few-shot examples, remains open \cite{wu-etal-2025-please}.

\paragraph{Limitations of reference-based metrics.}
Reference-based metrics like COMET \cite{guerreiro-etal-2024-xcomet} and MetricX \cite{juraska-etal-2024-metricx} primarily evaluate semantic adequacy, which can make them fail to capture cultural adaptation \cite{yuan2026cultureawaremt}. They also depend on a single reference, which can be one valid translation among many \cite{zouhar-etal-2024-pitfalls}. In the latest WMT meta-evaluation, \citet{lavie-etal-2025-findings} find that on harder test sets, ChrF \cite{popovic-2015-chrf} and LLM-as-judge correlate more strongly with human ratings than COMET or MetricX, calling into question the recent reliance on neural MT metrics. Further, LLM-based evaluation can integrate contextual information that these metrics do not account for \cite{sun2025finegrainedmultidimensionalmetricsdocumentlevel}, making it a more versatile way to evaluate translation.

\begin{figure}[t]
    \centering
    \includegraphics[width=\columnwidth]{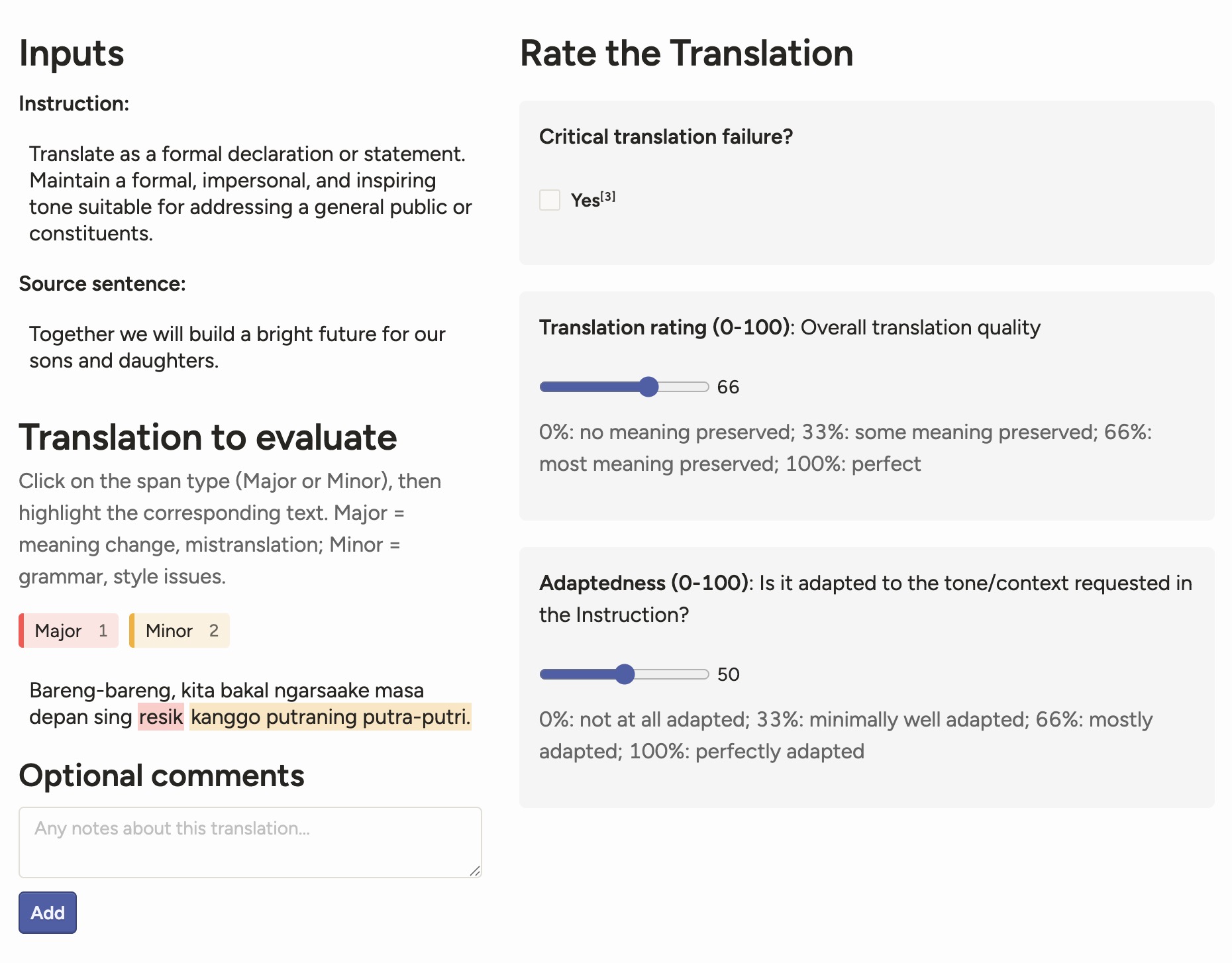}
    \caption{Annotation interface in Label Studio. Annotators see the instruction, source sentence, and model translation (left). They mark error spans and give a translation rating (ESA methodology), plus an adaptedness score (right).}
    \label{fig:annotation}
\end{figure}

\section{Methodology}
\label{sec:method}

\subsection{Task}

We study translation conditioned on both source text and a natural-language instruction describing the translation's intended purpose. Given source text $s$ and instruction $c$ (specifying audience, formality, domain, or other contextual information), the model produces translation $t = f(s, c)$. We compare this to the baseline $t_0 = f(s)$ where no instruction is provided. Figure~\ref{fig:pipeline} illustrates the full pipeline, with examples in Table~\ref{tab:examples}.

\subsection{Data}
\label{sec:data}

\paragraph{Dataset and languages.}
We evaluate on two benchmarks. BOUQuET \cite{andrews-etal-2025-bouquet} is our primary benchmark; we additionally use WMT24++ \cite{deutsch-etal-2025-wmt24} as a secondary, large-scale validation, verifying that our findings generalise beyond BOUQuET across over 50~languages (Table~\ref{tab:wmt24pp-domain} and Appendix~\ref{app:wmt24pp}).

For BOUQuET, we use the dev set (504~instances per language), which offers broad domain coverage (social media, how-to articles, literature, etc.)\ and diverse source-original languages. Each row includes metadata on domain, register, and comments describing context and intent, which we use for instruction generation (see below). All experiments are in the en$\rightarrow$xx direction, which is both in higher demand for deployed MT systems \cite{merx-etal-2025-low} and the harder direction for models \cite{zhu-etal-2024-multilingual}.
We run the full study, including human annotation, on five target languages: Indonesian, French, Ukrainian, Khmer, and Javanese. They span three scripts, three language families, and range from high- to low-resource. They also differ in how they grammaticalise register: French and Ukrainian express formality largely through a binary pronoun distinction, whereas Javanese and Khmer rely on more elaborate speech-level and honorific systems. This gives the set good coverage of register phenomena that instructions are meant to control.

\begin{figure*}[!htbp]
    \centering
    \includegraphics[width=\textwidth]{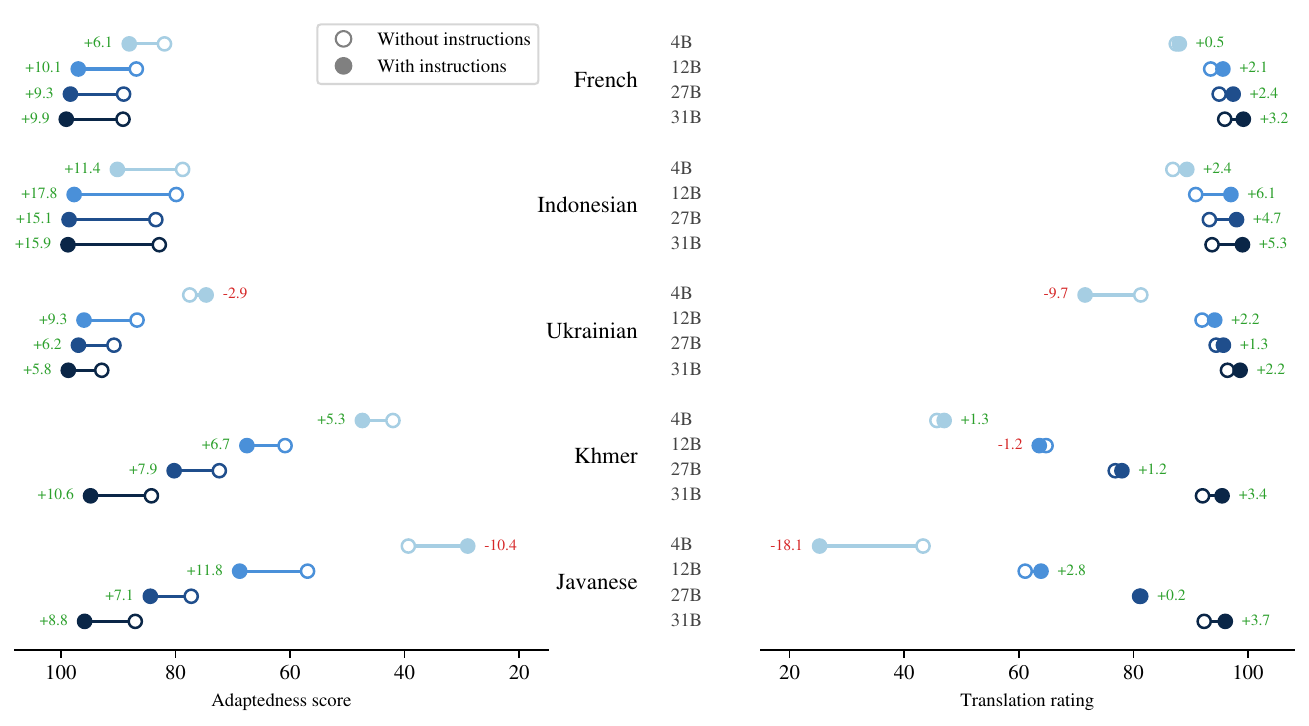}
    \caption{Effect of adding user instructions on adaptedness (left) and translation rating (right). Hollow circles = without instructions; filled circles = with instructions. Larger models get consistent improvements, while the 4B model can suffer from the presence of instructions. Note that 4B-27B are Gemma-3, 31B is Gemma-4.}
    \label{fig:dumbbell}
\end{figure*}

\begin{table}[t]
\centering
\small
\begin{adjustbox}{max width=\columnwidth}
\begin{tabular}{ll ccc cc}
\toprule
 & & \multicolumn{3}{c}{\textbf{LLM judge}} & \textbf{COMET} & \textbf{ChrF++} \\
\cmidrule(lr){3-5}
\textbf{Lang} & \textbf{Metric} & \textbf{$\alpha$} & \textbf{ICC} & \textbf{$r$} & \textbf{$r$} & \textbf{$r$} \\
\midrule
\multirow{2}{*}{fra}
 & Trans.\ Rating  & \cellcolor{win75}.862 & \cellcolor{win75}.928 & \cellcolor{win75}.867 & \cellcolor{win55}.562 & \cellcolor{win35}.360 \\
 & Adaptedness     & \cellcolor{win75}.817 & \cellcolor{win75}.901 & \cellcolor{win75}.820 & --    & --    \\
\midrule
\multirow{2}{*}{ind}
 & Trans.\ Rating  & \cellcolor{win75}.865 & \cellcolor{win75}.928 & \cellcolor{win75}.867 & \cellcolor{win35}.270 & \cellcolor{win45}.405 \\
 & Adaptedness     & \cellcolor{win55}.696 & \cellcolor{win75}.844 & \cellcolor{win65}.748 & --    & --    \\
\midrule
\multirow{2}{*}{ukr}
 & Trans.\ Rating  & \cellcolor{win65}.715 & \cellcolor{win75}.888 & \cellcolor{win65}.803 & \cellcolor{win55}.613 & \cellcolor{win45}.508 \\
 & Adaptedness     & \cellcolor{win45}.517 & \cellcolor{win65}.725 & \cellcolor{win55}.570 & --    & --    \\
\midrule
\multirow{2}{*}{khm}
 & Trans.\ Rating  & \cellcolor{win55}.637 & \cellcolor{win65}.783 & \cellcolor{win65}.731 & \cellcolor{win45}.513 & \cellcolor{win45}.499 \\
 & Adaptedness     & \cellcolor{win45}.509 & \cellcolor{win55}.675 & \cellcolor{win45}.535 & --    & --    \\
\midrule
\multirow{2}{*}{jav}
 & Trans.\ Rating  & \cellcolor{win65}.701 & \cellcolor{win65}.824 & \cellcolor{win65}.736 & \cellcolor{win55}.621 & \cellcolor{win35}.347 \\
 & Adaptedness     & \cellcolor{win55}.654 & \cellcolor{win65}.791 & \cellcolor{win55}.658 & --    & --    \\
\bottomrule
\end{tabular}
\end{adjustbox}
\caption{Human--model agreement on translation rating and adaptedness. $r$~= Pearson correlation; $\alpha$~= Krippendorff's alpha; ICC~= intraclass correlation ICC(3,$k$).}
\label{tab:human-llm-agreement}
\end{table}

\paragraph{Instruction generation.}
For each BOUQuET row, a natural-language instruction is drafted by Gemini~3~Flash (\citeauthor{geminiteam2025gemini3flash} \citeyear{geminiteam2025gemini3flash}, prompt in Appendix~\ref{app:prompt-instruction}), conditioning on the item's domain, register, and descriptive comment, then manually revised by a human annotator (Figure~\ref{fig:pipeline}(a)). Instructions describe the translation's intended audience, tone, and purpose, independently of target language (see examples in Table~\ref{tab:examples}).

\subsection{Translation}

\paragraph{Models} We evaluate three sizes of the Gemma~3 model family (\citeauthor{gemmateam2025gemma3technicalreport} \citeyear{gemmateam2025gemma3technicalreport} 4B, 12B, and 27B, instruction-tuned variants),\footnote{\href{https://huggingface.co/collections/google/gemma-3-release}{hf.co/collections/google/gemma-3-release}} alongside Gemma-4-31B.\footnote{\href{https://huggingface.co/google/gemma-4-31B}{hf.co/google/gemma-4-31B}}
We replicate our main findings on the Qwen3.5 model family \cite{qwenteam2026qwen35} in Appendix~\ref{app:qwen}.

\paragraph{Translation conditions.}
Each source sentence is translated under four conditions.
The \textbf{baseline} uses a fixed 3-shot prompt without instructions (Figure~\ref{fig:pipeline}, w/o line; prompt in \S\ref{app:prompt-translation-without-instructions}).
The \textbf{instruction} condition adds the user instruction to both the prompt and the 3-shot examples (Figure~\ref{fig:pipeline}, w/ line; prompt in \S\ref{app:prompt-translation-with-instructions}).
The \textbf{few-shot} condition replaces the fixed examples with 5~semantically similar translation pairs, retrieved from a held-out split (the BOUQuET test set), using the \texttt{all-MiniLM-L6-v2}\footnote{\href{https://huggingface.co/sentence-transformers/all-MiniLM-L6-v2}{hf.co/sentence-transformers/all-MiniLM-L6-v2}} embedding model.
The \textbf{paragraph} condition presents the full source paragraph, with markers delimiting the sentence to translate (Figure~\ref{fig:p1-p2-prompts}).
Translation is performed at the sentence level, except for the paragraph condition.

\paragraph{Inference.} All local models are served on a single A100 80GB GPU with vLLM using greedy decoding (temperature~0), an 8{,}192-token context window, and a 1{,}024-token generation cap, for a total of 100 GPU-hours across experiments. Qwen3.5 is run with reasoning disabled. All reported scores are means over the full item set (504 BOUQuET / 997 WMT24++ items per condition).

\subsection{Evaluation}
\label{sec:eval}

Our evaluation methodology measures translation rating (meaning preservation) and adaptedness (tone/context matching) through an LLM-judge, validated against human annotations.

\subsubsection{Translation rating and adaptedness}
\label{sec:human}

\paragraph{Translation rating.}
We score translations following the Error Span Annotation (ESA) protocol \citep{kocmi-etal-2024-error}, which is the standard for human translation evaluation as of WMT~2025 \citep{lavie-etal-2025-findings}. Annotators mark error spans (Minor / Major) in the translation and produce a 0--100 translation rating reflecting meaning preservation relative to the source (Figure~\ref{fig:annotation}).

\paragraph{Adaptedness.}
ESA measures meaning preservation, but a translation can preserve meaning yet fail on tone relative to audience \citep{house2015translation}, which instructions are designed to address. We therefore introduce adaptedness, a separate 0--100 score measuring how well the translation matches the instruction's intended purpose (scale guide shown to annotators in Figure~\ref{fig:annotation}). To validate that adaptedness captures a distinct construct, we collect human annotations from native speakers of each of the five target languages, 160~items per language. The two scores correlate strongly overall (Pearson $r=0.86$, pooled), but 12\% of items differ by more than 20 points on the 0--100 scale, and the divergence is concentrated in the conversation domain (21.4\% of items differ by $>$20 points).
Qualitatively, these disagreements are register failures invisible to meaning-based scoring: e.g., a Khmer translation rated 100 for meaning preservation but 50 for adaptedness, with the annotator noting ``Perfect translation, but the tone is too harsh''.

\begin{figure}[t]
    \centering
    \includegraphics[width=\columnwidth]{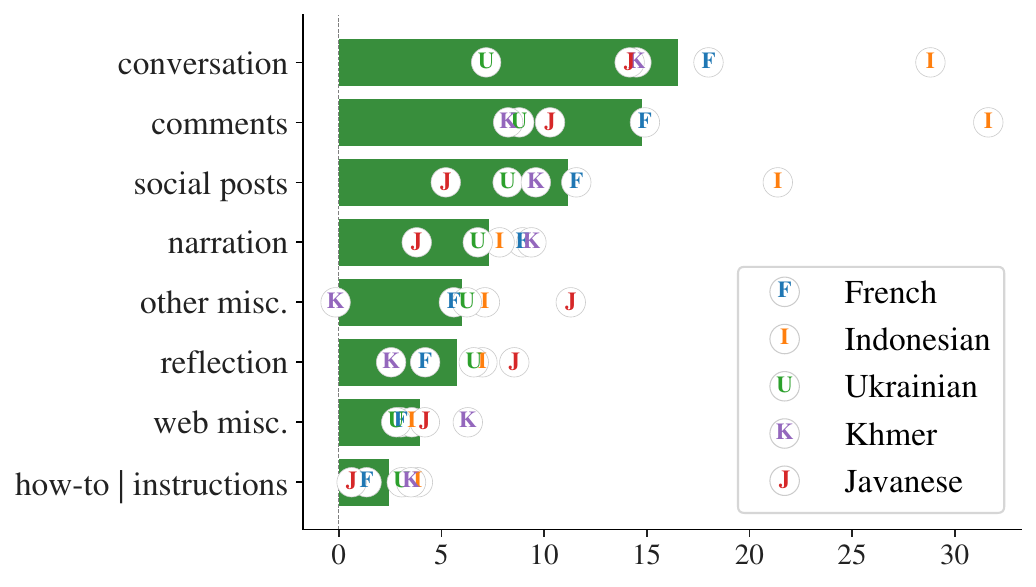}
    \caption{Average adaptedness $\Delta$ (with $-$ without instructions) per domain, for the 27B model.}
    \label{fig:domain}
\end{figure}

\subsubsection{Evaluation model}

\paragraph{Setup.}
To scale evaluation, we use an LLM judge (Gemini-3-Flash, full prompt in Appendix~\ref{app:prompt-judge}) that receives the source, instruction, and translation. It outputs (i)~a translation rating following ESA, (ii)~an adaptedness score, and (iii)~a critical-failure flag for output in the wrong language, or no translation at all. The judge always sees the instruction but is blind to whether it was present during translation. Appendix~\ref{app:judge-inputs} reports an ablation that separates the two criteria from the judge's inputs, scoring translation rating without the instruction and adaptedness without the source. Scoring is reference-free, consistent with ESA. Running the judge in reference-based mode, compared to reference-free, yields small, inconsistent differences in agreement with humans ($\Delta r \le 0.07$, $\Delta \alpha \le 0.07$), confirming that providing the reference does not materially improve LLM-judge performance. We also try Gemini-3.1-Pro in thinking mode, but find its agreement with humans is the same as Flash on translation rating (mean Pearson $r$ 0.80 for both, mean $\alpha$ 0.77 vs 0.76) and lower on adaptedness ($r$ 0.63 vs 0.67, $\alpha$ 0.58 vs 0.64), so we settle on Flash.

\paragraph{Validation.}
Table~\ref{tab:human-llm-agreement} reports LLM--human agreement on the native-speaker annotations. For all five languages, LLM--human agreement on translation rating is on par with or surpasses inter-annotator agreement typically reported for ESA (Pearson $r=0.48$; \citealt{kocmi-etal-2024-error}). We observe particularly strong LLM--human agreement for French and Indonesian (Pearson $r > 0.86$, $\alpha > 0.86$ for translation rating), with more moderate agreement for Khmer, Javanese, and Ukrainian ($r \ge 0.73$, $\alpha \ge 0.63$). COMET\footnote{XCOMET-XL: \href{https://huggingface.co/Unbabel/XCOMET-XL}{hf.co/Unbabel/XCOMET-XL}.} and ChrF++\footnote{Computed with sacrebleu~2.6.0; default parameters.} correlate substantially less with human judgement across all languages ($0.27 \le r \le 0.62$).

\section{Results}
\label{sec:results}

User instructions consistently improve translation quality (overall rating and adaptedness) across models and languages (Figure~\ref{fig:dumbbell}), with the exception of the 4B model on Ukrainian and Javanese, where critical failures degrade scores (discussed in \S\ref{sec:model-size}).

\begin{table}[t]
\centering
\small
\setlength{\tabcolsep}{3pt}
\begin{tabular}{l rrr rrr}
\toprule
& \multicolumn{3}{c}{\textbf{Adaptedness $\Delta$}} & \multicolumn{3}{c}{\textbf{Rating $\Delta$}} \\
\cmidrule(lr){2-4} \cmidrule(lr){5-7}
\textbf{Domain} & \textbf{fra} & \textbf{ind} & \textbf{ukr} & \textbf{fra} & \textbf{ind} & \textbf{ukr} \\
\midrule
social   & \cellcolor{dpos8}+21.0 & \cellcolor{dpos8}+26.1 & \cellcolor{dpos8}+19.9 & \cellcolor{dpos8}+11.6 & \cellcolor{dpos8}+13.4 & \cellcolor{dpos8}+10.0 \\
speech   & \cellcolor{dpos8}+12.4 & \cellcolor{dpos8}+16.2 & \cellcolor{dpos8}+11.6 & \cellcolor{dpos4}+6.4  & \cellcolor{dpos4}+7.5  & \cellcolor{dpos4}+6.1  \\
literary & \cellcolor{dpos4}+7.7  & \cellcolor{dpos8}+12.0 & \cellcolor{dpos8}+8.0  & \cellcolor{dpos0}+3.2  & \cellcolor{dpos4}+6.1  & \cellcolor{dpos0}+3.8  \\
news     & \cellcolor{dneg0}--1.1 & \cellcolor{dpos0}+2.2  & \cellcolor{dneg0}--0.2 & \cellcolor{dneg0}--1.5 & \cellcolor{dpos0}+0.9  & \cellcolor{dneg0}--0.2 \\
\bottomrule
\end{tabular}
\caption{Per-domain effect of adding instructions, WMT24++ 27B model ($\Delta$~=~with~$-$~without instructions). Informal domains (social, speech) show the largest gains in both adaptedness and translation rating across all languages, consistent with BOUQuET.}
\label{tab:wmt24pp-domain}
\end{table}

\subsection{Effect of model size}
\label{sec:model-size}

\textbf{Larger models benefit more.} The 12/27/31B models show consistent improvement when instructions are present across languages, both for adaptedness (+10.2) and translation rating (+2.6, Figure~\ref{fig:dumbbell}).
The 4B model has mixed results: It benefits from instructions on French and Indonesian (+6.1 / +11.4 adaptedness), but instruction-induced errors reduce scores for Ukrainian (-2.9) and Javanese (-10.4).

\textbf{Critical failure to follow instructions for smaller models.} For all languages, instructions make the 4B model more likely to produce critical failures (translation in the wrong language or no translation at all), raising the failure rate from 2.4--7.5\% (no instructions) to 4.8--45.5\% (with instructions); the largest jumps are on Javanese (+38.5~pts) and Ukrainian (+13.9~pts). The 12B and 27B models remain robust ($\le$0.6\% in either condition). Inspection reveals that these failures are predominantly outputs in the wrong language (e.g. Indonesian instead of Javanese), or a failure to perform the task (e.g. responding to a question instead of translating it), suggesting that the added complexity of processing contextual information overwhelms the smaller model's limited multilingual capacity.

Appendix~\ref{app:lora-sft} shows that \textbf{4B critical errors when instructions are present can largely be removed through LoRA distillation} \citep{hu2021lora}. Using the 27B model as teacher, we train a 4B adapter on SMOL \citep{caswell-etal-2025-smol}. The rate drops from 16.5\% to 0.0\% on Ukrainian and from 45.5\% to 4.0\% on Javanese (BOUQuET dev). The effect transfers across languages: an adapter trained only on Ukrainian lowers Javanese critical errors (45.5\% $\rightarrow$ 7.6\%), and does not reduce the model's ability to translate without instructions.

\subsection{Effect of language resource level}
\label{sec:language}

Ordering languages by approximate resource level (French~$>$~Indonesian~$>$~Ukrainian~$>$~Khmer~$>$~Javanese), win-rate results in Table~\ref{tab:winrate} show that \textbf{the higher the language resource level, the more models benefit from instructions}. This is validated by absolute score gains (Figure~\ref{fig:dumbbell}), with a caveat: when the baseline is already high (translation score $\geq$ 90), benchmark saturation means lower absolute gains (e.g. French 27B, 95 $\rightarrow$ 97).

Out of \textbf{50+~languages in WMT24++, for all but one, Gemma-3-27B benefits from the presence of instructions} (average +5.7 translation rating, +16.9 adaptedness; Figure~\ref{fig:wmt24pp-dumbbell}, Appendix~\ref{app:wmt24pp}). Aside from resource level, register richness (the degree to which a language grammaticalises social distinctions, e.g.\ Japanese keigo, Arabic address forms) also plays a role. High-resource register-rich languages (Arabic, Japanese) gain most (+20--28 adaptedness), ahead of high-resource register-light languages (French, Italian, Dutch; +13--16), while resource level still acts as a floor: very-low-resource languages (Gujarati, Zulu) benefit little even when register-rich.

\begin{table}[t]
\centering
\small
\setlength{\tabcolsep}{3pt}
\begin{adjustbox}{max width=\columnwidth}
\begin{tabular}{l ccccc}
\toprule
& \textbf{fra} & \textbf{ind} & \textbf{ukr} & \textbf{khm} & \textbf{jav} \\
\midrule
G3 4B  & \cellcolor{win55}54/5/40 & \cellcolor{win65}69/9/22 & \cellcolor{win45}48/5/48 & \cellcolor{win45}46/5/49 & \cellcolor{win35}38/1/61 \\
G3 12B & \cellcolor{win75}71/9/19 & \cellcolor{win75}75/11/14 & \cellcolor{win65}63/10/27 & \cellcolor{win55}52/2/46 & \cellcolor{win55}55/4/42 \\
G3 27B & \cellcolor{win75}72/12/16 & \cellcolor{win75}73/14/13 & \cellcolor{win65}69/8/23 & \cellcolor{win55}57/4/40 & \cellcolor{win65}63/3/34 \\
G4 31B & \cellcolor{win65}64/18/18 & \cellcolor{win75}73/13/14 & \cellcolor{win65}63/17/20 & \cellcolor{win65}67/8/25 & \cellcolor{win65}64/10/26 \\
\bottomrule
\end{tabular}
\end{adjustbox}
\caption{Win/tie/loss (\%) for instruction-conditioned translations vs.\ baseline in pairwise LLM-judge comparison. The larger the model, and the higher resource the language, the more likely the instruction-conditioned translation wins.}
\label{tab:winrate}
\end{table}

\subsection{Effect of domain}
\label{sec:domain}

\paragraph{Informal domains drive largest gains.}
Across languages, informal domains such as \textit{conversation} (+16.5) and \textit{social media comments} (+14.8) show the largest gains, while more formal domains like \textit{web misc.} (+4.0) and \textit{how-to/instructions} (+2.5) show smaller improvements (Figure~\ref{fig:domain}).

This domain-dependent pattern replicates on WMT24++ (Table~\ref{tab:wmt24pp-domain}): across 997~segments for French, Indonesian, and Ukrainian, the \textit{social} domain consistently shows the largest adaptedness gains (up to +26.1 for the 27B model on Indonesian), while \textit{news} shows negligible change.

\begin{table}[t]
\centering
\small
\setlength{\tabcolsep}{3.5pt}
\begin{tabular}{l rrrrr r}
\toprule
& \textbf{fra} & \textbf{ind} & \textbf{ukr} & \textbf{khm} & \textbf{jav} & \textbf{Avg.} \\
\midrule
Baseline & 86.8 & 79.9 & 86.7 & 60.9 & 56.9 & 74.2 \\
Few-shot & 88.4 & 84.1 & 89.5 & 61.6 & 62.1 & 77.1 \\
Paragraph & 84.8 & 78.6 & 83.8 & 55.6 & 56.0 & 71.8 \\
\textbf{Instr.} & \textbf{97.0} & \textbf{97.7} & \textbf{96.0} & \textbf{67.5} & \textbf{68.8} & \textbf{85.4} \\
\midrule
$\Delta$ FS - Base & \cellcolor{dpos0}+1.6 & \cellcolor{dpos4}+4.2 & \cellcolor{dpos0}+2.8 & \cellcolor{dpos0}+0.8 & \cellcolor{dpos4}+5.2 & \cellcolor{dpos0}+2.9 \\
$\Delta$ Para - Base & \cellcolor{dneg0}--2.1 & \cellcolor{dneg0}--1.2 & \cellcolor{dneg0}--2.9 & \cellcolor{dneg4}--5.2 & \cellcolor{dneg0}--0.9 & \cellcolor{dneg0}--2.5 \\
$\Delta$ Instr - Base & \cellcolor{dpos8}+10.1 & \cellcolor{dpos8}+17.8 & \cellcolor{dpos8}+9.3 & \cellcolor{dpos4}+6.7 & \cellcolor{dpos8}+11.8 & \cellcolor{dpos8}+11.1 \\
\bottomrule
\end{tabular}
\caption{Adaptedness for the 12B model. Baseline: sentence-level without instructions; Few-shot (FS): 5~semantically selected examples; Paragraph: sentence-in-context translation; Instructions: sentence-level with user instructions. Instructions provide the largest gains.}
\label{tab:fewshot-12b}
\end{table}

\begin{table}[t]
\centering
\small
\setlength{\tabcolsep}{1pt}
\begin{tabular}{@{}l@{\hskip 3pt}r@{\hskip 2pt}r@{\hskip 6pt}r@{\hskip 2pt}r@{\hskip 6pt}r@{\hskip 2pt}r@{}}
\toprule
& \multicolumn{2}{c}{\textbf{fra}} & \multicolumn{2}{c}{\textbf{ind}} & \multicolumn{2}{c}{\textbf{ukr}} \\
\cmidrule(lr){2-3} \cmidrule(lr){4-5} \cmidrule(lr){6-7}
\textbf{Domain} & \textbf{Div\%} & \textbf{$\Delta$C} & \textbf{Div\%} & \textbf{$\Delta$C} & \textbf{Div\%} & \textbf{$\Delta$C} \\
\midrule
conversation    & \cellcolor{win35}27 & \cellcolor{dneg4}$-$.028 & \cellcolor{win35}47 & \cellcolor{dneg4}$-$.028 & \cellcolor{win45}21 & \cellcolor{dneg4}$-$.039 \\
social comments        & \cellcolor{win35}33 & \cellcolor{dneg4}$-$.024 & \cellcolor{win35}50 & \cellcolor{dneg4}$-$.025 & \cellcolor{win35}29 & \cellcolor{dneg4}$-$.041 \\
social posts    & \cellcolor{win45}23 & \cellcolor{dneg0}$-$.012 & \cellcolor{win35}36 & \cellcolor{dneg4}$-$.036 & \cellcolor{win45}21 & \cellcolor{dneg4}$-$.031 \\
\midrule
how-to          & \cellcolor{win55} 6 & \cellcolor{dpos0}$+$.020 & \cellcolor{win55}13 & \cellcolor{dneg0}$-$.003 & \cellcolor{win55} 6 & \cellcolor{dpos0}$+$.003 \\
web misc.       & \cellcolor{win55} 6 & \cellcolor{dneg0}$-$.003 & \cellcolor{win55} 9 & \cellcolor{dneg0}$-$.012 & \cellcolor{win55}12 & \cellcolor{dpos0}$+$.001 \\
reflection pieces & \cellcolor{win55} 5 & \cellcolor{dpos0}$+$.019 & \cellcolor{win45}21 & \cellcolor{dneg0}$-$.006 & \cellcolor{win45}18 & \cellcolor{dneg0}$-$.020 \\
\bottomrule
\end{tabular}
\caption{LLM judge--COMET divergence (27B). Div\%: rows where the LLM judge prefers the instructed translation but COMET prefers the uninstructed one. $\Delta$C: mean XCOMET-XL change when instructions are added. Informal domains show the highest divergence.}
\label{tab:comet-domain}
\end{table}

\subsection{Comparison with semantic few-shot and paragraph-level translation}
\label{sec:fewshot}

Instructions yield substantially larger adaptedness gains than \textbf{semantic few-shot examples} across all 5~languages. For the 12B model (Table~\ref{tab:fewshot-12b}), the few-shot advantage over baseline averages +2.9~adaptedness points, vs +11.1 for explicit instructions. The pattern holds for the 27B model as well (few-shot +1.2; instruction +9.1, Table~\ref{tab:full-27b}).
For the 4B model, the pattern is consistent for higher-resource languages, but instructions-related critical failures on Ukrainian and Javanese mean that semantic few-shot is a more reliable strategy (Table~\ref{tab:full-4b}).

\textbf{Paragraph-level translation} is even less effective than semantic few-shot, and can even hurt translation quality.
We test two prompting strategies: $P_1$, which marks a single sentence for translation within the paragraph, and $P_2$, which translates all sentences at once in a numbered list.
$P_2$ consistently outperforms $P_1$, but its adaptedness gains over the baseline remain near zero for Gemma-3-27B (+0.3) and Qwen3.5-27B (+0.8), with modest gains for Gemma-4-31B (+1.5).
This is consistent with prior work finding that document-level context does not reliably improve translation quality for models $\leq$ 27B \cite{merx-etal-2025-openwho}.
Full paragraph-level setup and results can be found in Figure~\ref{fig:p1-p2-prompts} and Table~\ref{tab:paragraph-comparison}, Appendix~\ref{app:paragraph}.

Overall, for models $\geq$~12B, \textbf{explicit user instructions are both more reliable and more effective than semantic few-shot examples or paragraph-level context} for adapting translations to their intended purpose.
We return to this observation in \S\ref{sec:self-instr}: while paragraph context hurts when fed as raw input, it can be leveraged by the model to draft an instruction brief (Figure~\ref{fig:self-instr-pipeline}).

\subsection{Comparison with traditional MT metrics}
\label{sec:comet-divergence}

We find that XCOMET-XL and ChrF++ scores decrease when instructions are added, even as LLM-as-judge scores increase: XCOMET-XL drops by .014--.049 across all five languages, and ChrF++ by 3.7--11.1 points. Domain-level analysis reveals that this divergence is concentrated in informal domains: conversation, social media comments, and social posts show judge-COMET disagreement rates of 21--50\%, while formal domains (how-to, web misc.) have disagreement rates closer to 10\% (Table~\ref{tab:comet-domain}).
Qualitatively, we observe both a lower COMET performance on informal domains (better but more informal translations are rated lower), and the presence of rather formal BOUQuET reference translations, which may introduce reference bias.
Overall, our results are coherent with recent research finding that MT metrics can be inadequate for domains that are under-represented in their training distribution \cite{zouhar-etal-2024-pitfalls}, and on the sensitivity of reference-based metrics to reference quality \cite{lavie-etal-2025-findings}.

\section{Self-Instructing from Document Context}
\label{sec:self-instr}

\begin{figure}[t]
\centering
\begin{tikzpicture}[
    boxbase/.style={draw, rounded corners, minimum height=0.55cm, minimum width=1.4cm, font=\scriptsize, align=center, inner sep=2pt},
    metabox/.style={boxbase, fill=cmetafill, draw=cmetastroke},
    srcbox/.style={boxbase, fill=csrcfill, draw=csrcstroke},
    instrbox/.style={boxbase, fill=cinstrfill, draw=cinstrstroke},
    transbox/.style={boxbase, fill=ctransfill, draw=ctransstroke},
    llm/.style={draw=cllmstroke, circle, fill=white, minimum size=0.85cm, font=\scriptsize, align=center, inner sep=1pt},
    arr/.style={-{Stealth[length=3pt]}, thin},
    stagelbl/.style={font=\scriptsize\itshape, anchor=south},
]
\node[stagelbl, anchor=south west] at (-0.7, 0.825) {1. Generate instruction};
\node[metabox] (para) at (0, 0.55) {Paragraph\\context};
\node[llm] (genllm) at (2.3, 0.55) {LLM};
\node[instrbox] (selfinst) at (4.5, 0.55) {Self-\\instruction};

\node[stagelbl, anchor=south west] at (-0.7, -0.725) {2. Translate};
\node[srcbox] (sent) at (0, -1.0) {Sentence};
\node[llm] (transllm) at (2.3, -1.0) {LLM};
\node[transbox] (pred) at (4.5, -1.0) {Translated\\sentence};

\draw[arr] (para.east) -- (genllm.west);
\draw[arr] (genllm.east) -- (selfinst.west);
\draw[arr] (sent.east) -- (transllm.west);
\draw[arr] (transllm.east) -- (pred.west);
\draw[arr] (selfinst.south) |- (1.5,-0.1) |- (transllm.north west);
\end{tikzpicture}
\caption{Self-instruction pipeline. Gemma-4-31B first drafts a user instruction from the English paragraph and target sentence, then translates the sentence conditioned on its own generated instruction.}
\label{fig:self-instr-pipeline}
\end{figure}
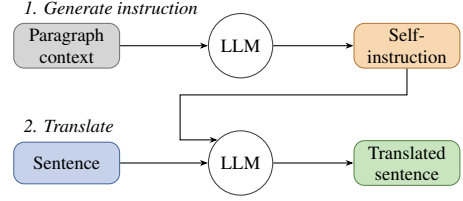

Throughout the main results, we use curated user instructions generated from document metadata (Figure~\ref{fig:pipeline}a). In practice, users may not write such briefs: translation systems receive a source sentence, at best with its surrounding document context. This raises a practical question: can a model ``guess'' its own instruction from surrounding context, recovering the benefit of a curated brief?

Two observations motivate this.
First, feeding paragraph context to the LLM as raw text can hurt adaptedness (-2.3 for Gemma-3~27B, \S\ref{sec:fewshot}).
Second, an instruction ablation (Appendix~\ref{app:ablation}) shows that most of the adaptedness benefit of a curated instruction comes from its \textit{context} component ("this is a conversation between friends") rather than \textit{explicit directives} ("use an informal tone").
Together, they suggest that externalising paragraph context as an instruction may help adapt translation to context.

\subsection{Pipeline}

Figure~\ref{fig:self-instr-pipeline} illustrates the pipeline: for each sentence, we prompt Gemma-\{3-27B,4-31B\} with the English source paragraph, marking the target sentence with a \texttt{[TARGET]} tag, and ask the model to return a concise user instruction specifying the inferred setting, target audience, and tone. We use a 3-shot prompt with greedy decoding. The resulting \textbf{Self-instruction} replaces the gold one in the translation pipeline, with every other setting held fixed.

\begin{table}[t]
\centering
\small
\setlength{\tabcolsep}{4pt}
\begin{tabular}{l rr rr}
\toprule
 & \multicolumn{2}{c}{\textbf{Gemma-3-27B}} & \multicolumn{2}{c}{\textbf{Gemma-4-31B}} \\
\cmidrule(lr){2-3} \cmidrule(lr){4-5}
 & \textbf{Adapt.} & \textbf{Rating} & \textbf{Adapt.} & \textbf{Rating} \\
\midrule
$\Delta\;P_1-$Base       & \cellcolor{dneg0}--2.3 & \cellcolor{dneg0}--3.1 & \cellcolor{dpos0}+0.9 & \cellcolor{dpos0}+0.3 \\
$\Delta\;P_2-$Base       & \cellcolor{dpos0}+0.3 & \cellcolor{dpos0}+0.2 & \cellcolor{dpos0}+1.5 & \cellcolor{dpos0}+0.6 \\
$\Delta$ Self-Instr$-$Base & \cellcolor{dpos4}+6.5 & \cellcolor{dpos0}+0.8 & \cellcolor{dpos8}+8.2 & \cellcolor{dpos0}+2.7 \\
$\Delta$ Curated$-$Base    & \cellcolor{dpos8}+9.1 & \cellcolor{dpos0}+2.0 & \cellcolor{dpos8}+10.2 & \cellcolor{dpos0}+3.6 \\
\bottomrule
\end{tabular}
\caption{Mean adaptedness and translation rating $\Delta$ vs.\ sentence-level baseline, averaged across the five languages. Self-Instr captures most of the Curated-instruction benefit (71\% on 27B, 80\% on 31B for adaptedness). For Gemma-3-27B, externalising paragraph context can flip it from harmful (--2.3 with $P_1$) to beneficial (+6.5). Per-language results in Appendix~\ref{app:self-instr-gemma3}.}
\label{tab:self-instr-summary}
\end{table}

\subsection{Results}

\textbf{Self-generated instructions close most of the curated-instruction gap.}
Across models, Self-Instr closes most of the adaptedness gap (71\% on 27B, 80\% on 31B), and much of the translation-rating gap (43\% on 27B, 77\% on 31B).
The effect is positive for every language (Appendix~\ref{app:self-instr-gemma3}).

\textbf{Externalising paragraph context can flip it from harmful to beneficial.}
Raw paragraph context \textit{decreases} Gemma-3-27B adaptedness by --2.3 on average ($P_1$, \S\ref{sec:fewshot}); the same paragraph, converted into a brief through Self-Instr, yields large adaptedness gains (+6.5), even though both settings give the model the same inputs (source sentence and paragraph context).
Qualitatively, by inspecting Gemma-4-31B thinking traces, we find that LLMs tend to produce a translation draft immediately and then deliberate over individual word choices, rather than considering target audience and appropriate register.

On \textbf{WMT24++} (three languages, longer documents), \textbf{Self-instruction replicates}, closing 74\% of the adaptedness gap with Gemma-3-27B, while raw document context hurts adaptedness (-10.6) (Appendix~\ref{app:self-instr-gemma3}, Table~\ref{tab:self-instr-wmt-27b}).

Overall, paragraph context, once externalised as an instruction, improves translation adaptedness to purpose, albeit to a lesser extent than curated (gold) instructions, which benefit from access to document metadata, and from a stronger drafting pipeline (Gemini~Flash + human post-editing).

\section{Discussion}
\label{sec:discussion}

\paragraph{Purpose-adapted MT is a measurable capability.}
Purpose-adapted MT treats translation as faithful to its communicative purpose and audience, not only to the surface of the source text. Our results show that current LLMs can meaningfully act on this broader notion of fidelity: given an explicit instruction, they produce measurably better-adapted translations. The effect is strongest on informal domains such as conversation and social media.

\paragraph{Need for diverse domains in evaluation.}
Traditional MT evaluation suites such as FLORES \cite{goyal-etal-2022-flores} are built from Wikipedia and news articles, close to the general-purpose text LLMs are pre-trained on, and under-represent conversational text, where the adaptation gap is largest. A plausible mechanism is that LLMs default to the written register of their pretraining distribution; instructions help most on conversational data because they nudge the model off that default. 
Comprehensive MT evaluation therefore requires diverse domains, including conversational text.

\paragraph{The instruction-failure tradeoff.}
For smaller models on low-resource languages, instructions introduce a tradeoff: added complexity triggers catastrophic failures, even though the model can adapt when it does produce target-language output. The 4B model on Javanese illustrates this: 45.5\% critical failure rate with instructions (vs.~7.0\% without), yet among non-failed translations, the adaptedness delta is +14.7. This suggests that \textbf{instruction-following and multilingual generation compete for limited model capacity}, and the instruction can ``crowd out'' the translation skill. Appendix~\ref{app:lora-sft} shows that PEFT distillation from a 27B teacher removes this tradeoff, reducing 4B instruction-conditioned failures on Javanese from 45.5\% to 4.0\%.

\paragraph{Practical implications.}
For models $\geq$~12B parameters, adding user instructions is a reliable way to improve translation quality (\S\ref{sec:results}). When users do not supply an instruction, self-instruction from surrounding document context (\S\ref{sec:self-instr}) offers a practical recipe: the model drafts its own brief before translating, recovering most of the adaptedness benefit of a curated instruction. The strong performance of instruction-based MT over few-shot examples (\S\ref{sec:fewshot}) shows that for most languages, models offer a natural way to follow user intent without the need for an external database of examples.

\section{Conclusion}
\label{sec:conclusion}

We study purpose-informed translation, where users supply instructions about target audience and communicative intent alongside the source text, and ask whether current LLMs can act on that specification. Across 50+~languages, 5~model sizes, and 8~domains, they can: instructions produce substantial gains in adaptedness, largest on informal domains and scaling with both model size and language resource level.

Two takeaways follow. First, translation evaluation should condition on purpose (contrary to traditional MT metrics). Second, when no instruction is supplied, a capable model can draft one from surrounding document context (self-instruction) and recover most of the adaptedness benefit of a curated brief, giving practitioners a concrete deployment path for document-level MT.

Looking forward, LLMs have opened new translation capabilities that are more adaptive and iterative \cite{liu2025newtrendsmodernmachine}. Without evaluation to measure these capabilities, we risk an over-emphasis on rigid translation flows (one input, one output). Better measuring these capabilities will give us a basis to improve them.

\section*{Limitations}

\paragraph{Limited few-shot comparison.}
Our few-shot comparison (\S\ref{sec:fewshot}) uses a single retrieval setup: 5~examples retrieved via cosine similarity over \texttt{all-MiniLM-L6-v2} embeddings from a held-out BOUQuET split. A more thorough comparison would vary the embedding model, retrieval method (e.g., BM25, multilingual embedders), and example count. Our finding that instructions outperform semantically-matched few-shot examples is therefore a claim about one widespread few-shot configuration, and is not necessarily exhaustive.

\paragraph{Limited language coverage.}
We focus on five target languages spanning three scripts and a range of resource levels, but none from Africa and none very-low-resource. However, our WMT24++ validation (Appendix~\ref{app:wmt24pp}) expands to over 50 languages, where the main findings replicate (Figure~\ref{fig:wmt24pp-dumbbell}). The controllable-MT benchmarks in Appendix~\ref{app:controlled-mt} cover eight further languages each, but only French overlaps our five in-depth languages, so those replications and our human validation rest on largely disjoint language sets.
Similarly, we only evaluate with source text in English, which as a very register-light language, might benefit more from explicit instructions than register-rich languages, which carry more information about target audience in the source text. We likewise do not evaluate language pairs with English on neither side, leaving this to future work.

\paragraph{Limited model diversity.}
Main results rest on Gemma models, evaluated by a Gemini-3-Flash judge. All three share a vendor ecosystem, so we cannot rule out shared biases between the translator and the judge. Appendix~\ref{app:qwen} replicates the core findings on Qwen3.5, including the COMET-judge divergence and the informal-domain pattern, which reduces this concern. Broader replication across open-weight families would strengthen the scope of our claims.

\paragraph{LLM-as-judge reliability.}
Our LLM--human agreement on translation rating ranges from Pearson $r$=0.73 on Khmer to $r$=0.87 on French (Table~\ref{tab:human-llm-agreement}), on par with typical inter-annotator agreement for ESA \citep{kocmi-etal-2024-error}. Adaptedness agreement is lower and uneven across languages, particularly on Khmer ($r$=0.54) and Ukrainian ($r$=0.57), so findings on those languages should be read as more exploratory than for French or Indonesian.

\section*{Ethical Considerations}

\paragraph{Human annotators.}
Translation evaluation annotations (\S\ref{sec:eval}) were collected from one native-speaker annotator per target language: French, Indonesian, Ukrainian, Khmer, and a Javanese--Indonesian bilingual annotator for Javanese.
Annotators were recruited through the authors' personal networks and compensated at approximately US\$20/hour, above the national minimum wage in the relevant jurisdictions and above prevailing local rates for similar annotation work in the annotators' countries of residence.
Informed consent was obtained prior to annotation.
The protocol was not subject to formal ethics review: annotation consists of rating short benchmark sentences with no collection of personal data and no exposure to risky content. Further protocol details are in Appendix~\ref{app:annotation-protocol}.

\paragraph{Risks.}
We propose an evaluation and prompting framework; it does not introduce translation capabilities beyond those of the underlying models. Instruction-conditioned translation could in principle be used to register-shift text for impersonation or persuasion, but this risk applies to any controllable-style generation system.

\paragraph{Artifact use.}
We use BOUQuET~\citep{andrews-etal-2025-bouquet}, WMT24++~\citep{deutsch-etal-2025-wmt24}, SMOL~\citep{caswell-etal-2025-smol}, CoCoA-MT~\citep[CDLA-Sharing-1.0]{nadejde-etal-2022-cocoa}, MT-GenEval~\citep[CC-BY-SA-3.0]{currey-etal-2022-mt}, Gemma-3 / Gemma-4~\citep{gemmateam2025gemma3technicalreport}, Qwen3.5~\citep{qwenteam2026qwen35}, NLLB-3.3B~\citep{nllb-2022-no}, and all-MiniLM-L6-v2 within their published license terms; all permit research use. BOUQuET and WMT24++ are curated translation benchmarks vetted for content by their creators; we did not encounter personally-identifying or offensive material in the subsets evaluated. Model outputs were spot-checked during annotation.

\paragraph{AI assistant disclosure.}
LLM-based coding assistants were used for code authoring and analysis, and LLMs were used for proofreading the manuscript. All claims, experimental designs, and analysis decisions are the authors'.

\section*{Acknowledgements}
We gratefully acknowledge the work of our language experts: Artem Fedorinchyk, Tungga Dewi, Victor Ilen, and Rith Pit.
This research was supported by an Australian Research Council Discovery Early Career Researcher Award (DECRA project number DE260100695) held by Ekaterina Vylomova.
This research was supported by The University of Melbourne's Research Computing Services and the Petascale Campus.

\bibliography{anthology,custom}

\appendix

\section{Full results per language}
\label{app:ratings}

\paragraph{Translation evaluation.} We report translation rating and adaptedness for all four models we evaluate (Gemma-3-4B in Table~\ref{tab:full-4b}, 12B in Table~\ref{tab:full-12b}, 27B in Table~\ref{tab:full-27b}, and Gemma-4-31B in Table~\ref{tab:full-gemma4-31b}) on BOUQuET dev. Each model is evaluated under four conditions: \textbf{Baseline} (sentence-level, no instruction), \textbf{Few-shot} (5~semantically retrieved examples), \textbf{Paragraph} (sentence-in-context, $P_1$ prompt), and \textbf{Instructions} (gold per-sentence brief).

\paragraph{Statistical significance.} All reported scores are means over the per-condition item set (504~items). We test the instruction effect ($\Delta$~Instr$-$Base) with a paired bootstrap over items (10{,}000 resamples). Averaged over the five languages, instructions change adaptedness by $+1.9$ (95\%~CI $[+0.5, +3.3]$) for 4B, $+11.1$ ($[+10.2, +12.1]$) for 12B, $+9.1$ ($[+8.3, +10.0]$) for 27B, and $+10.2$ ($[+9.5, +11.0]$) for Gemma-4-31B; the corresponding translation-rating changes are $-4.7$ ($[-6.1, -3.4]$), $+2.4$ ($[+1.7, +3.2]$), $+2.0$ ($[+1.4, +2.6]$), and $+3.6$ ($[+3.2, +4.0]$). For models $\geq$12B, every per-language adaptedness gain is significant (p$<$0.001); translation-rating gains are smaller and not always significant (e.g.\ Khmer for the 12B and 27B models, Javanese for 27B). For the 4B model, the Ukrainian and Javanese rating drops are significant (p$<$0.001), reflecting the critical-failure tradeoff (\S\ref{sec:model-size}).

\begin{table*}[!htbp]
\centering
\small
\begin{tabular}{l rrrrr r rrrrr r}
\toprule
& \multicolumn{6}{c}{\textbf{Translation Rating}} & \multicolumn{6}{c}{\textbf{Adaptedness}} \\
\cmidrule(lr){2-7} \cmidrule(lr){8-13}
& \textbf{fra} & \textbf{ind} & \textbf{ukr} & \textbf{khm} & \textbf{jav} & \textbf{Avg.} & \textbf{fra} & \textbf{ind} & \textbf{ukr} & \textbf{khm} & \textbf{jav} & \textbf{Avg.} \\
\midrule
Baseline & 87.5 & 86.9 & 81.3 & 45.7 & 43.3 & 68.9 & 81.9 & 78.7 & 77.5 & 42.0 & 39.3 & 63.9 \\
Few-shot & 88.2 & 88.5 & 84.3 & 45.2 & \textbf{46.6} & \textbf{70.6} & 83.0 & 81.8 & 80.4 & 41.8 & \textbf{43.2} & \textbf{66.1} \\
Paragraph & \textbf{91.7} & 88.8 & \textbf{86.5} & 38.0 & 39.6 & 68.9 & 84.1 & 78.9 & \textbf{82.0} & 34.5 & 35.7 & 63.1 \\
\textbf{Instructions} & 88.0 & \textbf{89.3} & 71.6 & \textbf{47.0} & 25.2 & 64.2 & \textbf{88.1} & \textbf{90.1} & 74.6 & \textbf{47.4} & 29.0 & 65.8 \\
\midrule
$\Delta$ FS$-$Base & \cellcolor{dpos0}+0.7 & \cellcolor{dpos0}+1.6 & \cellcolor{dpos4}+3.0 & \cellcolor{dneg0}--0.5 & \cellcolor{dpos4}+3.3 & \cellcolor{dpos0}+1.6 & \cellcolor{dpos0}+1.1 & \cellcolor{dpos4}+3.1 & \cellcolor{dpos0}+2.9 & \cellcolor{dneg0}--0.2 & \cellcolor{dpos4}+3.9 & \cellcolor{dpos0}+2.2 \\
$\Delta$ Para$-$Base & \cellcolor{dpos4}+4.2 & \cellcolor{dpos0}+1.9 & \cellcolor{dpos4}+5.2 & \cellcolor{dneg4}--7.8 & \cellcolor{dneg4}--3.7 & \cellcolor{dneg0}--0.0 & \cellcolor{dpos0}+2.2 & \cellcolor{dpos0}+0.2 & \cellcolor{dpos4}+4.5 & \cellcolor{dneg4}--7.5 & \cellcolor{dneg4}--3.6 & \cellcolor{dneg0}--0.8 \\
$\Delta$ Instr$-$Base & \cellcolor{dpos0}+0.5 & \cellcolor{dpos0}+2.4 & \cellcolor{dneg4}--9.7 & \cellcolor{dpos0}+1.3 & \cellcolor{dneg4}--18.1 & \cellcolor{dneg4}--4.7 & \cellcolor{dpos4}+6.1 & \cellcolor{dpos8}+11.4 & \cellcolor{dneg0}--2.9 & \cellcolor{dpos4}+5.3 & \cellcolor{dneg4}--10.4 & \cellcolor{dpos0}+1.9 \\
\bottomrule
\end{tabular}
\caption{Gemma-3-4B on BOUQuET dev (504~items per cell). Instruction-induced critical failures collapse the rating on Ukrainian and Javanese; few-shot is the most reliable strategy at this scale.}
\label{tab:full-4b}
\end{table*}

\begin{table*}[!htbp]
\centering
\small
\begin{tabular}{l rrrrr r rrrrr r}
\toprule
& \multicolumn{6}{c}{\textbf{Translation Rating}} & \multicolumn{6}{c}{\textbf{Adaptedness}} \\
\cmidrule(lr){2-7} \cmidrule(lr){8-13}
& \textbf{fra} & \textbf{ind} & \textbf{ukr} & \textbf{khm} & \textbf{jav} & \textbf{Avg.} & \textbf{fra} & \textbf{ind} & \textbf{ukr} & \textbf{khm} & \textbf{jav} & \textbf{Avg.} \\
\midrule
Baseline & 93.5 & 90.9 & 92.0 & \textbf{64.8} & 61.1 & 80.5 & 86.8 & 79.9 & 86.7 & 60.9 & 56.9 & 74.2 \\
Few-shot & 93.2 & 92.3 & 93.1 & 64.3 & 63.4 & 81.3 & 88.4 & 84.1 & 89.5 & 61.6 & 62.1 & 77.1 \\
Paragraph & 91.4 & 89.0 & 89.4 & 59.1 & 60.7 & 77.9 & 84.8 & 78.6 & 83.8 & 55.6 & 56.0 & 71.8 \\
\textbf{Instructions} & \textbf{95.6} & \textbf{97.0} & \textbf{94.2} & 63.6 & \textbf{63.9} & \textbf{82.9} & \textbf{97.0} & \textbf{97.7} & \textbf{96.0} & \textbf{67.5} & \textbf{68.8} & \textbf{85.4} \\
\midrule
$\Delta$ FS$-$Base & \cellcolor{dneg0}--0.3 & \cellcolor{dpos0}+1.5 & \cellcolor{dpos0}+1.1 & \cellcolor{dneg0}--0.5 & \cellcolor{dpos0}+2.3 & \cellcolor{dpos0}+0.8 & \cellcolor{dpos0}+1.6 & \cellcolor{dpos4}+4.2 & \cellcolor{dpos0}+2.8 & \cellcolor{dpos0}+0.8 & \cellcolor{dpos4}+5.2 & \cellcolor{dpos0}+2.9 \\
$\Delta$ Para$-$Base & \cellcolor{dneg0}--2.1 & \cellcolor{dneg0}--1.9 & \cellcolor{dneg0}--2.6 & \cellcolor{dneg4}--5.7 & \cellcolor{dneg0}--0.4 & \cellcolor{dneg0}--2.5 & \cellcolor{dneg0}--2.1 & \cellcolor{dneg0}--1.2 & \cellcolor{dneg0}--2.9 & \cellcolor{dneg4}--5.2 & \cellcolor{dneg0}--0.9 & \cellcolor{dneg0}--2.5 \\
$\Delta$ Instr$-$Base & \cellcolor{dpos0}+2.1 & \cellcolor{dpos4}+6.1 & \cellcolor{dpos0}+2.2 & \cellcolor{dneg0}--1.2 & \cellcolor{dpos0}+2.8 & \cellcolor{dpos0}+2.4 & \cellcolor{dpos8}+10.1 & \cellcolor{dpos8}+17.8 & \cellcolor{dpos8}+9.3 & \cellcolor{dpos4}+6.7 & \cellcolor{dpos8}+11.8 & \cellcolor{dpos8}+11.1 \\
\bottomrule
\end{tabular}
\caption{Gemma-3-12B on BOUQuET dev (504~items per cell). Instructions yield the largest gains across all five languages; paragraph context consistently hurts.}
\label{tab:full-12b}
\end{table*}

\begin{table*}[!htbp]
\centering
\small
\begin{tabular}{l rrrrr r rrrrr r}
\toprule
& \multicolumn{6}{c}{\textbf{Translation Rating}} & \multicolumn{6}{c}{\textbf{Adaptedness}} \\
\cmidrule(lr){2-7} \cmidrule(lr){8-13}
& \textbf{fra} & \textbf{ind} & \textbf{ukr} & \textbf{khm} & \textbf{jav} & \textbf{Avg.} & \textbf{fra} & \textbf{ind} & \textbf{ukr} & \textbf{khm} & \textbf{jav} & \textbf{Avg.} \\
\midrule
Baseline & 95.0 & 93.3 & 94.5 & 76.8 & 81.1 & 88.1 & 89.0 & 83.4 & 90.7 & 72.3 & 77.2 & 82.6 \\
Few-shot & 96.0 & 93.9 & 95.1 & 77.5 & \textbf{81.6} & 88.8 & 91.6 & 86.0 & 91.9 & 72.0 & 77.4 & 83.8 \\
Paragraph & 92.9 & 90.7 & 91.5 & 72.9 & 77.3 & 85.0 & 88.2 & 81.9 & 89.0 & 68.2 & 74.2 & 80.3 \\
\textbf{Instructions} & \textbf{97.4} & \textbf{98.0} & \textbf{95.7} & \textbf{78.0} & 81.3 & \textbf{90.1} & \textbf{98.3} & \textbf{98.6} & \textbf{96.9} & \textbf{80.2} & \textbf{84.4} & \textbf{91.7} \\
\midrule
$\Delta$ FS$-$Base & \cellcolor{dpos0}+1.1 & \cellcolor{dpos0}+0.7 & \cellcolor{dpos0}+0.6 & \cellcolor{dpos0}+0.7 & \cellcolor{dpos0}+0.5 & \cellcolor{dpos0}+0.7 & \cellcolor{dpos0}+2.6 & \cellcolor{dpos0}+2.6 & \cellcolor{dpos0}+1.2 & \cellcolor{dneg0}--0.4 & \cellcolor{dpos0}+0.2 & \cellcolor{dpos0}+1.2 \\
$\Delta$ Para$-$Base & \cellcolor{dneg0}--2.1 & \cellcolor{dneg0}--2.6 & \cellcolor{dneg4}--3.0 & \cellcolor{dneg4}--4.0 & \cellcolor{dneg4}--3.8 & \cellcolor{dneg4}--3.1 & \cellcolor{dneg0}--0.9 & \cellcolor{dneg0}--1.6 & \cellcolor{dneg0}--1.7 & \cellcolor{dneg4}--4.1 & \cellcolor{dneg4}--3.0 & \cellcolor{dneg0}--2.3 \\
$\Delta$ Instr$-$Base & \cellcolor{dpos0}+2.4 & \cellcolor{dpos4}+4.7 & \cellcolor{dpos0}+1.3 & \cellcolor{dpos0}+1.2 & \cellcolor{dpos0}+0.2 & \cellcolor{dpos0}+2.0 & \cellcolor{dpos8}+9.3 & \cellcolor{dpos8}+15.1 & \cellcolor{dpos4}+6.2 & \cellcolor{dpos4}+7.9 & \cellcolor{dpos4}+7.1 & \cellcolor{dpos8}+9.1 \\
\bottomrule
\end{tabular}
\caption{Gemma-3-27B on BOUQuET dev (504~items per cell). Instructions outperform all alternatives on adaptedness across the board, and on rating in every language except Javanese, where few-shot edges them out by 0.3.}
\label{tab:full-27b}
\end{table*}

\begin{table*}[!htbp]
\centering
\small
\begin{tabular}{l rrrrr r rrrrr r}
\toprule
& \multicolumn{6}{c}{\textbf{Translation Rating}} & \multicolumn{6}{c}{\textbf{Adaptedness}} \\
\cmidrule(lr){2-7} \cmidrule(lr){8-13}
& \textbf{fra} & \textbf{ind} & \textbf{ukr} & \textbf{khm} & \textbf{jav} & \textbf{Avg.} & \textbf{fra} & \textbf{ind} & \textbf{ukr} & \textbf{khm} & \textbf{jav} & \textbf{Avg.} \\
\midrule
Baseline & 96.0 & 93.7 & 96.4 & 92.1 & 92.4 & 94.1 & 89.1 & 82.8 & 92.9 & 84.2 & 87.0 & 87.2 \\
Few-shot & 97.0 & 94.8 & 96.9 & 92.2 & 92.9 & 94.7 & 92.1 & 85.6 & 93.4 & 84.7 & 87.2 & 88.6 \\
Paragraph & 96.1 & 94.0 & 96.6 & 92.8 & 92.4 & 94.4 & 90.3 & 83.8 & 93.4 & 85.4 & 87.4 & 88.1 \\
\textbf{Instructions} & \textbf{99.2} & \textbf{99.0} & \textbf{98.6} & \textbf{95.5} & \textbf{96.0} & \textbf{97.7} & \textbf{99.0} & \textbf{98.8} & \textbf{98.7} & \textbf{94.8} & \textbf{95.9} & \textbf{97.4} \\
\midrule
$\Delta$ FS$-$Base & \cellcolor{dpos0}+1.0 & \cellcolor{dpos0}+1.1 & \cellcolor{dpos0}+0.5 & \cellcolor{dpos0}+0.1 & \cellcolor{dpos0}+0.5 & \cellcolor{dpos0}+0.6 & \cellcolor{dpos4}+3.0 & \cellcolor{dpos0}+2.8 & \cellcolor{dpos0}+0.5 & \cellcolor{dpos0}+0.5 & \cellcolor{dpos0}+0.2 & \cellcolor{dpos0}+1.4 \\
$\Delta$ Para$-$Base & \cellcolor{dpos0}+0.2 & \cellcolor{dpos0}+0.3 & \cellcolor{dpos0}+0.2 & \cellcolor{dpos0}+0.7 & \cellcolor{dpos0}+0.0 & \cellcolor{dpos0}+0.3 & \cellcolor{dpos0}+1.2 & \cellcolor{dpos0}+1.0 & \cellcolor{dpos0}+0.5 & \cellcolor{dpos0}+1.2 & \cellcolor{dpos0}+0.4 & \cellcolor{dpos0}+0.9 \\
$\Delta$ Instr$-$Base & \cellcolor{dpos4}+3.2 & \cellcolor{dpos4}+5.3 & \cellcolor{dpos0}+2.2 & \cellcolor{dpos4}+3.4 & \cellcolor{dpos4}+3.7 & \cellcolor{dpos4}+3.6 & \cellcolor{dpos8}+9.9 & \cellcolor{dpos8}+15.9 & \cellcolor{dpos4}+5.8 & \cellcolor{dpos8}+10.6 & \cellcolor{dpos8}+8.8 & \cellcolor{dpos8}+10.2 \\
\bottomrule
\end{tabular}
\caption{Gemma-4-31B on BOUQuET dev (504~items per cell). Instructions improve both rating and adaptedness across all five languages, with no critical-failure regression.}
\label{tab:full-gemma4-31b}
\end{table*}

\paragraph{Traditional MT metrics.} Table~\ref{tab:comet-full} provides XCOMET-XL and ChrF++ scores for all model sizes and instruction conditions.

\begin{table*}[!htbp]
\centering
\small
\begin{tabular}{ll ccccc ccccc}
\toprule
& & \multicolumn{5}{c}{\textbf{XCOMET-XL}} & \multicolumn{5}{c}{\textbf{ChrF++}} \\
\cmidrule(lr){3-7} \cmidrule(lr){8-12}
\textbf{Model} & \textbf{Instr.} & \textbf{fra} & \textbf{ind} & \textbf{ukr} & \textbf{khm} & \textbf{jav} & \textbf{fra} & \textbf{ind} & \textbf{ukr} & \textbf{khm} & \textbf{jav} \\
\midrule
\multirow{2}{*}{4B}
 & w/o & .862 & .891 & .908 & .595 & .625 & 62.4 & 57.2 & 52.3 & 26.4 & 35.2 \\
 & w/ & \cellcolor{dpos0}.874 & \cellcolor{dneg8}.836 & \cellcolor{dneg8}.858 & \cellcolor{dneg0}.576 & \cellcolor{dpos8}.707 & \cellcolor{dneg8}51.5 & \cellcolor{dneg8}46.1 & \cellcolor{dneg8}34.4 & \cellcolor{dneg0}25.2 & \cellcolor{dneg8}20.6 \\
\midrule
\multirow{2}{*}{12B}
 & w/o & .943 & .938 & .944 & .717 & .747 & 64.9 & 60.8 & 59.2 & 30.5 & 45.9 \\
 & w/ & \cellcolor{dneg4}.917 & \cellcolor{dneg0}.932 & \cellcolor{dneg4}.922 & \cellcolor{dneg8}.647 & \cellcolor{dneg4}.709 & \cellcolor{dneg4}58.0 & \cellcolor{dneg8}51.4 & \cellcolor{dneg8}47.9 & \cellcolor{dneg0}27.9 & \cellcolor{dneg8}38.3 \\
\midrule
\multirow{2}{*}{27B}
 & w/o & .938 & .947 & .951 & .796 & .832 & 66.3 & 61.9 & 60.8 & 38.5 & 52.0 \\
 & w/ & \cellcolor{dneg0}.924 & \cellcolor{dneg4}.919 & \cellcolor{dneg4}.924 & \cellcolor{dneg4}.755 & \cellcolor{dneg4}.783 & \cellcolor{dneg4}59.4 & \cellcolor{dneg8}53.3 & \cellcolor{dneg8}49.7 & \cellcolor{dneg4}34.8 & \cellcolor{dneg8}41.5 \\
\bottomrule
\end{tabular}
\caption{XCOMET-XL and ChrF++ scores across all models and languages. Cell shading on the \textbf{w/} rows encodes the delta versus the \textbf{w/o} baseline (red = decrease, green = increase; darker = larger magnitude). Reference-based metrics consistently decrease when instructions are added, in contrast to LLM-as-judge metrics that show improvement. The lone exception is Javanese on the 4B model, where XCOMET-XL rises (+.082) despite the sharp performance drop measured by the LLM-judge, presumably because XCOMET-XL fails to flag wrong-language output (Indonesian).}
\label{tab:comet-full}
\end{table*}

\section{Comparison with NLLB-3.3B}
\label{app:nllb}

To situate instruction-following MT against traditional NMT models, we compare Gemma-3 to NLLB-200-3.3B \cite{nllb-2022-no} on BOUQuET dev. NLLB does not accept instructions, so it is evaluated in a single, unconditioned setting. Translations are produced with beam search ($k$=4) and scored with the same LLM-as-judge protocol used throughout the paper (Gemini-3-Flash, both translation rating and adaptedness; the judge sees the gold instruction in all conditions).

Table~\ref{tab:nllb-deltas} reports Gemma-3 scores expressed as $\Delta$ versus NLLB-3.3B. The 4B model trails NLLB substantially on average ($-11.6$ rating, $-9.0$ adaptedness without instructions), and instructions widen the gap on Ukrainian and Javanese due to the critical-failure tradeoff discussed in \S\ref{sec:model-size}. The \textbf{12B model is roughly on par with NLLB without instructions} (Avg.\ $-0.0$ rating, $+1.3$ adaptedness) \textbf{and pulls clearly ahead with instructions} ($+2.4$ rating, $+12.5$ adaptedness). The 27B model already beats NLLB without instructions on every language, and instructions widen the lead to $+9.6$ rating / $+18.7$ adaptedness.

To better understand the 12B\,$\leftrightarrow$\,NLLB tradeoff, Table~\ref{tab:nllb-12b-domain} breaks Gemma-3-12B $-$ NLLB down by domain, pooling across the five languages.
\textbf{Without instructions, 12B trades wins and losses with NLLB by domain.} It already beats NLLB on conversation ($+5.9$ rating, $+11.6$ adaptedness) and matches it on most other domains, but loses on more formal text, e.g. web misc.\ ($-4.1$/$-4.5$), reflection ($-4.3$/$-5.9$). \textbf{With instructions, the 12B model is competitive or better than NLLB on every domain, and the adaptedness gap is concentrated in informal text}: conversation $+29.6$, social comments $+21.1$, social posts $+12.5$.

\begin{table*}[!htbp]
\centering
\small
\setlength{\tabcolsep}{4pt}
\begin{tabular}{l rrrrr r rrrrr r}
\toprule
& \multicolumn{6}{c}{\textbf{Translation Rating}} & \multicolumn{6}{c}{\textbf{Adaptedness}} \\
\cmidrule(lr){2-7} \cmidrule(lr){8-13}
& \textbf{fra} & \textbf{ind} & \textbf{ukr} & \textbf{khm} & \textbf{jav} & \textbf{Avg.} & \textbf{fra} & \textbf{ind} & \textbf{ukr} & \textbf{khm} & \textbf{jav} & \textbf{Avg.} \\
\midrule
NLLB-3.3B (raw) & 86.9 & 82.4 & 83.1 & 71.9 & 78.2 & 80.5 & 79.0 & 71.0 & 77.0 & 65.6 & 71.9 & 72.9 \\
\midrule
$\Delta$ Gemma-3-4B w/o
 & \cellcolor{dpos0}+0.6 & \cellcolor{dpos4}+4.5 & \cellcolor{dneg0}--1.8 & \cellcolor{dneg4}--26.1 & \cellcolor{dneg4}--34.9 & \cellcolor{dneg4}--11.6
 & \cellcolor{dpos0}+2.9 & \cellcolor{dpos4}+7.7 & \cellcolor{dpos0}+0.5 & \cellcolor{dneg4}--23.6 & \cellcolor{dneg4}--32.5 & \cellcolor{dneg4}--9.0 \\
$\Delta$ Gemma-3-4B w/
 & \cellcolor{dpos0}+1.1 & \cellcolor{dpos4}+6.9 & \cellcolor{dneg4}--11.6 & \cellcolor{dneg4}--24.9 & \cellcolor{dneg4}--52.9 & \cellcolor{dneg4}--16.3
 & \cellcolor{dpos8}+9.1 & \cellcolor{dpos8}+19.1 & \cellcolor{dneg0}--2.3 & \cellcolor{dneg4}--18.3 & \cellcolor{dneg4}--42.9 & \cellcolor{dneg4}--7.1 \\
\midrule
$\Delta$ Gemma-3-12B w/o
 & \cellcolor{dpos4}+6.6 & \cellcolor{dpos8}+8.5 & \cellcolor{dpos8}+8.9 & \cellcolor{dneg4}--7.1 & \cellcolor{dneg4}--17.0 & \cellcolor{dneg0}--0.0
 & \cellcolor{dpos4}+7.9 & \cellcolor{dpos8}+8.8 & \cellcolor{dpos8}+9.7 & \cellcolor{dneg4}--4.8 & \cellcolor{dneg4}--14.9 & \cellcolor{dpos0}+1.3 \\
$\Delta$ Gemma-3-12B w/
 & \cellcolor{dpos8}+8.7 & \cellcolor{dpos8}+14.6 & \cellcolor{dpos8}+11.1 & \cellcolor{dneg4}--8.3 & \cellcolor{dneg4}--14.3 & \cellcolor{dpos0}+2.4
 & \cellcolor{dpos8}+18.0 & \cellcolor{dpos8}+26.6 & \cellcolor{dpos8}+19.0 & \cellcolor{dpos0}+1.9 & \cellcolor{dneg0}--3.1 & \cellcolor{dpos8}+12.5 \\
\midrule
$\Delta$ Gemma-3-27B w/o
 & \cellcolor{dpos8}+8.0 & \cellcolor{dpos8}+10.9 & \cellcolor{dpos8}+11.3 & \cellcolor{dpos4}+5.0 & \cellcolor{dpos0}+2.9 & \cellcolor{dpos4}+7.6
 & \cellcolor{dpos8}+10.1 & \cellcolor{dpos8}+12.4 & \cellcolor{dpos8}+13.8 & \cellcolor{dpos4}+6.7 & \cellcolor{dpos4}+5.4 & \cellcolor{dpos8}+9.7 \\
$\Delta$ Gemma-3-27B w/
 & \cellcolor{dpos8}+10.3 & \cellcolor{dpos8}+15.8 & \cellcolor{dpos8}+12.7 & \cellcolor{dpos4}+5.6 & \cellcolor{dpos0}+3.4 & \cellcolor{dpos8}+9.6
 & \cellcolor{dpos8}+19.2 & \cellcolor{dpos8}+27.6 & \cellcolor{dpos8}+20.0 & \cellcolor{dpos8}+13.8 & \cellcolor{dpos8}+13.2 & \cellcolor{dpos8}+18.7 \\
\bottomrule
\end{tabular}
\caption{BOUQuET dev (504~items per language). Top row: NLLB-3.3B raw scores from the LLM-judge. Remaining rows: Gemma-3 scores expressed as $\Delta$ versus NLLB. The 12B model matches NLLB without instructions (Avg.\ $\approx 0$ rating) and overtakes it on adaptedness once instructions are added; the 27B model beats NLLB across the board. The 4B model falls below NLLB, especially on lower-resource languages.}
\label{tab:nllb-deltas}
\end{table*}

\begin{table*}[!htbp]
\centering
\small
\setlength{\tabcolsep}{5pt}
\begin{tabular}{l r rrr rrr}
\toprule
& & \multicolumn{3}{c}{\textbf{Translation Rating}} & \multicolumn{3}{c}{\textbf{Adaptedness}} \\
\cmidrule(lr){3-5} \cmidrule(lr){6-8}
\textbf{Domain} & \textbf{n} & \textbf{NLLB} & \textbf{$\Delta$ w/o} & \textbf{$\Delta$ w/} & \textbf{NLLB} & \textbf{$\Delta$ w/o} & \textbf{$\Delta$ w/} \\
\midrule
conversation          & 470  & 72.6 & \cellcolor{dpos4}+5.9 & \cellcolor{dpos8}+8.6 & 54.4 & \cellcolor{dpos8}+11.6 & \cellcolor{dpos8}+29.6 \\
narration             & 375  & 77.2 & \cellcolor{dneg0}--0.1 & \cellcolor{dpos0}+0.9 & 68.3 & \cellcolor{dpos0}+1.4  & \cellcolor{dpos8}+8.7  \\
how-to / instructions & 350  & 79.0 & \cellcolor{dpos0}+1.4 & \cellcolor{dpos0}+2.5 & 82.0 & \cellcolor{dneg0}--0.2 & \cellcolor{dpos4}+4.1  \\
social posts          & 350  & 83.1 & \cellcolor{dneg0}--2.0 & \cellcolor{dpos0}+0.5 & 76.0 & \cellcolor{dneg0}--1.1 & \cellcolor{dpos8}+12.5 \\
web misc.             & 330  & 88.4 & \cellcolor{dneg4}--4.1 & \cellcolor{dneg0}--0.4 & 86.8 & \cellcolor{dneg4}--4.5 & \cellcolor{dpos0}+2.5  \\
comments              & 290  & 81.7 & \cellcolor{dpos0}+0.6 & \cellcolor{dpos0}+3.7 & 69.0 & \cellcolor{dpos0}+3.7  & \cellcolor{dpos8}+21.1 \\
reflection            & 190  & 83.9 & \cellcolor{dneg4}--4.3 & \cellcolor{dneg0}--0.6 & 79.9 & \cellcolor{dneg4}--5.9 & \cellcolor{dpos4}+4.9  \\
other misc.           & 165  & 86.3 & \cellcolor{dneg0}--3.7 & \cellcolor{dneg0}--1.7 & 81.2 & \cellcolor{dneg0}--3.7 & \cellcolor{dpos4}+4.0  \\
\midrule
ALL                   & 2520 & 80.5 & \cellcolor{dneg0}--0.0 & \cellcolor{dpos0}+2.3 & 72.9 & \cellcolor{dpos0}+1.3  & \cellcolor{dpos8}+12.5 \\
\bottomrule
\end{tabular}
\caption{Per-domain comparison of Gemma-3-12B against NLLB-3.3B, pooled across all five languages ($n$ = pooled item count per domain). NLLB columns show its raw rating / adaptedness; $\Delta$ columns show Gemma-3-12B $-$ NLLB without and with user instructions. Without instructions, 12B trades wins and losses with NLLB by domain (losing on formal-reference text, winning on conversation). With instructions, 12B matches or beats NLLB everywhere, with the adaptedness advantage concentrated in informal domains.}
\label{tab:nllb-12b-domain}
\end{table*}

\section{WMT24++ Validation}
\label{app:wmt24pp}

To verify that our findings generalise beyond BOUQuET, we run the same LLM-judge pipeline on WMT24++ \cite{deutsch-etal-2025-wmt24}, which has 997~segments per language across 4~domains (social, literary, news, speech).

\paragraph{Instruction generation.}
Compared to BOUQuET, WMT24++ does not carry as much per-segment metadata (register, contextual comments), so instructions are obtained solely from the domain label and the source (English) document context.
We generate instructions with Gemini~3~Flash, using a 3-shot prompt (Appendix~\ref{app:prompt-instruction}). Resulting instructions describe the translation setting, target audience, and tone in 1--3~sentences, similar to examples in Table~\ref{tab:examples}. Instructions are target-language-agnostic (one instruction per English source segment, reused across every WMT24++ target language). Unlike BOUQuET, these instructions are not human post-edited.

\textbf{Per-domain breakdown} is reported in the main body (Table~\ref{tab:wmt24pp-domain}): similar to BOUQuET results, informal domains (social and speech) benefit most from the presence of explicit instructions (+17.9 adaptedness on average), while literary (+9.2) and news (+0.3) show more moderate and uneven gains.

\begin{figure*}[!htbp]
\centering
\includegraphics[width=\textwidth]{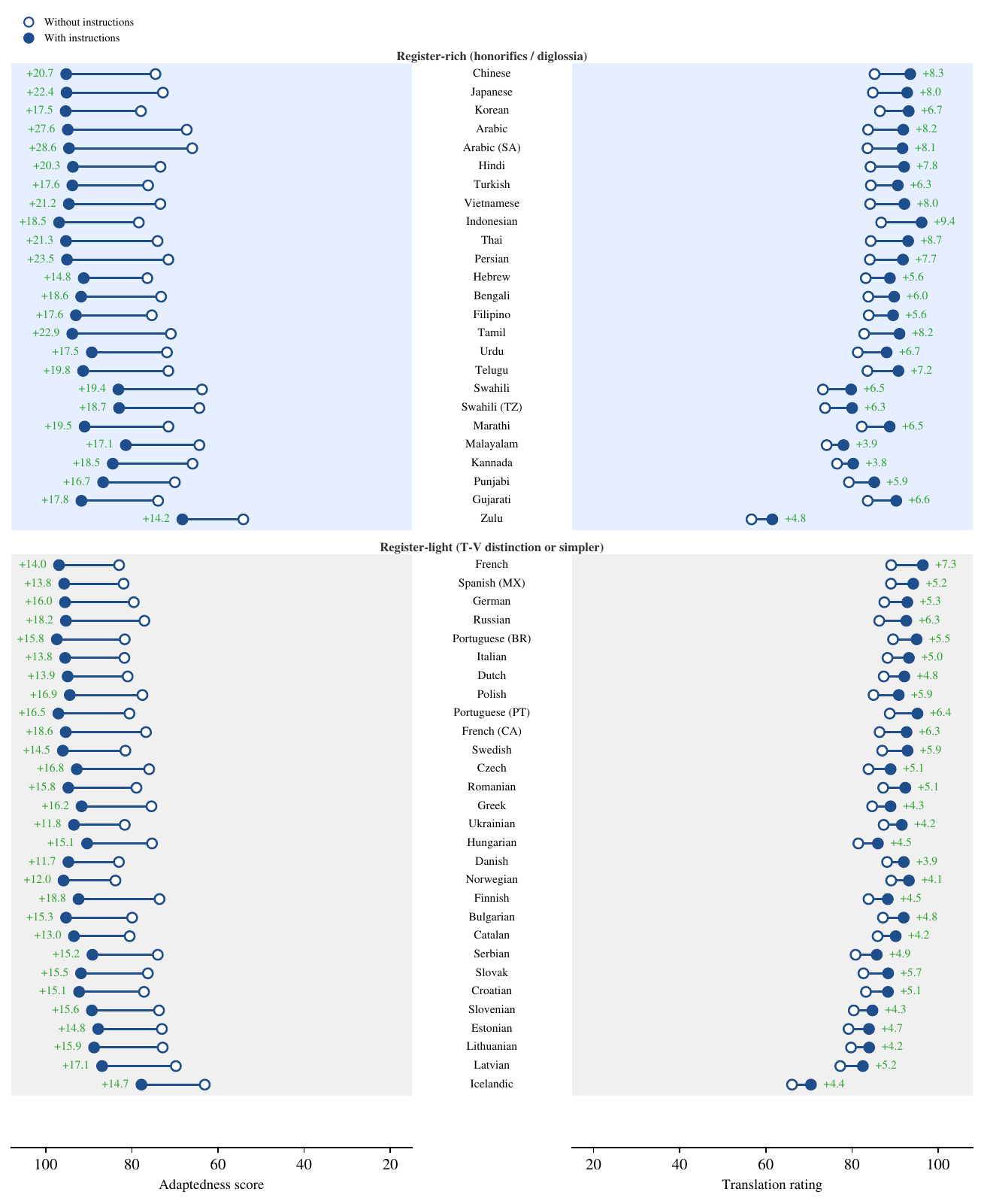}
\caption{WMT24++ results for Gemma-3-27B (997 segments per language), grouped by register richness: the upper band has languages with grammaticalised honorifics/register morphology, the lower band has languages with simple to no registers (mostly European languages). Within each band, languages are ordered by resource level (highest at the top). The BOUQuET trends replicate: the model consistently benefits from instructions, more so on register-rich languages.}
\label{fig:wmt24pp-dumbbell}
\end{figure*}

\paragraph{Expansion to 50+ languages.} Figure~\ref{fig:wmt24pp-dumbbell} broadens the picture to over 50~WMT24++ languages with Gemma-3-27B. 
\textbf{Resource richness} seems to determine how much a language benefits from explicit instructions.
Register-rich languages (Arabic, Japanese) show the largest adaptedness gains (+20--28), ahead of register-light languages (French, Italian, Dutch; +13--16).

\begin{table*}[!htbp]
\centering
\small
\setlength{\tabcolsep}{3pt}
\begin{tabular}{l rrrrr r rrrrr r}
\toprule
 & \multicolumn{6}{c}{\textbf{Translation Rating}} & \multicolumn{6}{c}{\textbf{Adaptedness}} \\
\cmidrule(lr){2-7} \cmidrule(lr){8-13}
 & \textbf{fra} & \textbf{ind} & \textbf{ukr} & \textbf{zul} & \textbf{guj} & \textbf{Avg.} & \textbf{fra} & \textbf{ind} & \textbf{ukr} & \textbf{zul} & \textbf{guj} & \textbf{Avg.} \\
\midrule
Gemma-3-4B & \cellcolor{dneg8}--57.6 & \cellcolor{dneg8}--21.1 & \cellcolor{dneg8}--26.0 & \cellcolor{dneg4}--2.3 & \cellcolor{dneg8}--36.9 & --28.8 & \cellcolor{dneg8}--49.7 & \cellcolor{dneg8}--10.3 & \cellcolor{dneg8}--18.9 & \cellcolor{dpos0}+0.3 & \cellcolor{dneg8}--29.9 & --21.7 \\
Gemma-3-12B & \cellcolor{dpos8}+7.3 & \cellcolor{dpos8}+10.5 & \cellcolor{dpos8}+6.7 & \cellcolor{dpos4}+4.1 & \cellcolor{dpos4}+5.7 & +6.8 & \cellcolor{dpos8}+14.9 & \cellcolor{dpos8}+20.5 & \cellcolor{dpos8}+13.9 & \cellcolor{dpos8}+12.2 & \cellcolor{dpos8}+15.3 & +15.4 \\
Gemma-3-27B & \cellcolor{dpos8}+7.3 & \cellcolor{dpos8}+9.4 & \cellcolor{dpos8}+6.8 & \cellcolor{dpos4}+4.8 & \cellcolor{dpos8}+6.6 & +7.0 & \cellcolor{dpos8}+14.0 & \cellcolor{dpos8}+18.5 & \cellcolor{dpos8}+13.5 & \cellcolor{dpos8}+14.2 & \cellcolor{dpos8}+17.8 & +15.6 \\
\bottomrule
\end{tabular}
\caption{Effect of user instructions on WMT24++ translation rating and adaptedness ($\Delta$ = Instr $-$ Base) across Gemma-3 model scales, for the five languages where all three model sizes were evaluated. The 4B model collapses under instructions.}
\label{tab:wmt24pp-gemma-scale}
\end{table*}

\paragraph{Comparison with document-level translation}
\label{para:wmt-doc-translation}
Table~\ref{tab:full-wmt-27b} reports the full Baseline / Paragraph / Instructions breakdown for Gemma-3-27B on French, Indonesian, and Ukrainian, comparing both paragraph prompting strategies from \S\ref{sec:fewshot}: $P_1$ (single sentence marked within its document) and $P_2$ (whole document translated at once as a numbered list). The pattern from BOUQuET (Table~\ref{tab:full-27b}) replicates and is amplified: explicit instructions improve translation rating by +7.8 and adaptedness by +15.3 on average, while paragraph context $P_1$ \textit{hurts} both (--7.0 rating, --10.6 adaptedness). The $P_1$ regression is larger than on BOUQuET, presumably because WMT24++ documents are longer than BOUQuET paragraphs. As on BOUQuET, $P_2$ consistently outperforms $P_1$: it nearly recovers baseline translation rating (--1.1), but adaptedness still lags the baseline by --6.7. Neither paragraph strategy approaches the gains from explicit instructions.

\begin{table*}[!htbp]
\centering
\small
\begin{tabular}{l rrr r rrr r}
\toprule
& \multicolumn{4}{c}{\textbf{Translation Rating}} & \multicolumn{4}{c}{\textbf{Adaptedness}} \\
\cmidrule(lr){2-5} \cmidrule(lr){6-9}
& \textbf{fra} & \textbf{ind} & \textbf{ukr} & \textbf{Avg.} & \textbf{fra} & \textbf{ind} & \textbf{ukr} & \textbf{Avg.} \\
\midrule
Baseline & 89.1 & 86.8 & 87.4 & 87.8 & 83.0 & 78.4 & 81.6 & 81.0 \\
Paragraph ($P_1$) & 82.3 & 79.2 & 80.6 & 80.7 & 72.5 & 65.0 & 73.8 & 70.4 \\
Paragraph ($P_2$) & 88.3 & 84.9 & 86.7 & 86.6 & 77.2 & 68.9 & 76.7 & 74.3 \\
\textbf{Instructions} & \textbf{96.5} & \textbf{96.2} & \textbf{94.1} & \textbf{95.6} & \textbf{96.9} & \textbf{96.9} & \textbf{95.2} & \textbf{96.3} \\
\midrule
$\Delta\;P_1-$Base & \cellcolor{dneg8}--6.8 & \cellcolor{dneg8}--7.6 & \cellcolor{dneg8}--6.7 & \cellcolor{dneg8}--7.0 & \cellcolor{dneg8}--10.4 & \cellcolor{dneg8}--13.4 & \cellcolor{dneg8}--7.8 & \cellcolor{dneg8}--10.6 \\
$\Delta\;P_2-$Base & \cellcolor{dneg0}--0.8 & \cellcolor{dneg0}--1.9 & \cellcolor{dneg0}--0.7 & \cellcolor{dneg0}--1.1 & \cellcolor{dneg8}--5.8 & \cellcolor{dneg8}--9.5 & \cellcolor{dneg4}--4.9 & \cellcolor{dneg8}--6.7 \\
$\Delta$ Instr$-$Base & \cellcolor{dpos8}+7.3 & \cellcolor{dpos8}+9.4 & \cellcolor{dpos8}+6.8 & \cellcolor{dpos8}+7.8 & \cellcolor{dpos8}+14.0 & \cellcolor{dpos8}+18.5 & \cellcolor{dpos8}+13.5 & \cellcolor{dpos8}+15.3 \\
\bottomrule
\end{tabular}
\caption{Gemma-3-27B on WMT24++ (997 items per cell). Baseline is the 3-shot sentence-level prompt without instructions; Paragraph~$P_1$ wraps the target sentence in [START\_TRANSLATE]\dots[END\_TRANSLATE] markers within its full source document; Paragraph~$P_2$ translates the whole document at once as a numbered list; Instructions adds the gold per-sentence brief. Instructions help across the board; both paragraph strategies hurt adaptedness, $P_2$ less than $P_1$, neither approaching the instruction gains.}
\label{tab:full-wmt-27b}
\end{table*}

\clearpage

\setlength{\textfloatsep}{6pt plus 2pt minus 2pt}
\setlength{\floatsep}{6pt plus 2pt minus 2pt}
\setlength{\intextsep}{6pt plus 2pt minus 2pt}
\renewcommand{\arraystretch}{0.88}
\raggedbottom

\section{Qwen 3.5 Validation}
\label{app:qwen}

\begin{table*}[t]
\centering
\small
\begin{tabular}{l rrrrr r rrrrr r}
\toprule
& \multicolumn{6}{c}{\textbf{Translation Rating}} & \multicolumn{6}{c}{\textbf{Adaptedness}} \\
\cmidrule(lr){2-7} \cmidrule(lr){8-13}
& \textbf{fra} & \textbf{ind} & \textbf{ukr} & \textbf{khm} & \textbf{jav} & \textbf{Avg.} & \textbf{fra} & \textbf{ind} & \textbf{ukr} & \textbf{khm} & \textbf{jav} & \textbf{Avg.} \\
\midrule
Baseline & 94.0 & 92.4 & 93.1 & 80.1 & 80.3 & 88.0 & 86.1 & 80.7 & 88.9 & 72.8 & 75.1 & 80.7 \\
Few-shot & 94.9 & 92.4 & 93.0 & 79.9 & 80.4 & 88.1 & 88.6 & 82.0 & 89.0 & 72.4 & 75.2 & 81.4 \\
Paragraph & 94.5 & 92.3 & 93.2 & 79.8 & \textbf{80.5} & 88.1 & 87.3 & 80.6 & 88.6 & 73.1 & 75.3 & 81.0 \\
\textbf{Instructions} & \textbf{98.0} & \textbf{97.5} & \textbf{95.0} & \textbf{82.9} & 78.3 & \textbf{90.3} & \textbf{97.6} & \textbf{97.4} & \textbf{96.0} & \textbf{81.8} & \textbf{79.4} & \textbf{90.4} \\
\midrule
$\Delta$ FS$-$Base & \cellcolor{dpos0}+0.9 & \cellcolor{dpos0}+0.0 & \cellcolor{dneg0}--0.1 & \cellcolor{dneg0}--0.2 & \cellcolor{dpos0}+0.1 & +0.1 & \cellcolor{dpos0}+2.5 & \cellcolor{dpos0}+1.3 & \cellcolor{dpos0}+0.1 & \cellcolor{dneg0}--0.4 & \cellcolor{dpos0}+0.1 & +0.7 \\
$\Delta$ Para$-$Base & \cellcolor{dpos0}+0.5 & \cellcolor{dneg0}--0.1 & \cellcolor{dpos0}+0.1 & \cellcolor{dneg0}--0.3 & \cellcolor{dpos0}+0.2 & +0.1 & \cellcolor{dpos0}+1.2 & \cellcolor{dneg0}--0.1 & \cellcolor{dneg0}--0.3 & \cellcolor{dpos0}+0.3 & \cellcolor{dpos0}+0.2 & +0.3 \\
$\Delta$ Instr$-$Base & \cellcolor{dpos0}+4.0 & \cellcolor{dpos4}+5.1 & \cellcolor{dpos0}+1.9 & \cellcolor{dpos0}+2.8 & \cellcolor{dneg0}--2.0 & +2.4 & \cellcolor{dpos8}+11.5 & \cellcolor{dpos8}+16.7 & \cellcolor{dpos4}+7.1 & \cellcolor{dpos8}+9.0 & \cellcolor{dpos4}+4.3 & +9.7 \\
\bottomrule
\end{tabular}
\caption{Translation rating and adaptedness scores for the Qwen3.5-27B model: baseline, semantic few-shot (5~examples), sentence-in-context paragraph translation, and instructions. Paragraph context yields near-zero deltas, consistent with the Gemma-3-27B results.}
\label{tab:fewshot-qwen-27b}
\end{table*}

To verify that our findings are not specific to the Gemma model family, we \textbf{replicate the core experiments on Qwen3.5 (4B, 9B, 27B)}.

The core findings replicate: (1)~instructions improve adaptedness by +9.7 points on average (Table~\ref{tab:fewshot-qwen-27b}); (2)~larger models show higher win rates (73.7\%~$\rightarrow$~81.5\%, Table~\ref{tab:qwen-main}); and (3)~XCOMET-XL decreases with instructions despite improved adaptedness (Table~\ref{tab:qwen-comet}), confirming the metric divergence observed with Gemma. The domain pattern also replicates (Table~\ref{tab:qwen-domain}): informal domains (comments, conversation) show the largest gains, while formal domains (web misc., how-to) show the smallest.

\begin{table}[!htbp]
\centering
\small
\setlength{\tabcolsep}{4pt}
\resizebox{\columnwidth}{!}{%
\begin{tabular}{l rr rr r}
\toprule
& \multicolumn{2}{c}{\textbf{Adapt.}} & \multicolumn{2}{c}{\textbf{Err. \%}} & \textbf{Win} \\
\cmidrule(lr){2-3} \cmidrule(lr){4-5}
\textbf{Model} & w/o & w/ & w/o & w/ & \textbf{Rate} \\
\midrule
Qwen3.5-4B  & 78.8 & 91.8 \goodup{13.0} & 0.2 & 0.0 & 73.7\% \\
Qwen3.5-9B  & 81.0 & 95.5 \goodup{14.5} & 0.0 & 0.0 & 79.0\% \\
Qwen3.5-27B & 83.4 & 97.6 \goodup{14.2} & 0.0 & 0.0 & 81.5\% \\
\bottomrule
\end{tabular}%
}
\caption{Qwen3.5 on Indonesian (BOUQuET dev, 504~items). Win rate = \% of pairwise comparisons where the instructed translation is preferred. The Gemma patterns hold: instructions improve adaptedness, larger models benefit more.}
\label{tab:qwen-main}
\end{table}

\begin{table}[!htbp]
\centering
\small
\begin{tabular}{l cc}
\toprule
& \multicolumn{2}{c}{\textbf{XCOMET-XL}} \\
\cmidrule(lr){2-3}
\textbf{Model} & w/o & w/ \\
\midrule
Qwen3.5-4B  & .941 & .923 \\
Qwen3.5-9B  & .948 & .939 \\
Qwen3.5-27B & .954 & .951 \\
\bottomrule
\end{tabular}
\caption{XCOMET-XL for Qwen3.5 on Indonesian. As with Gemma, COMET decreases when instructions are added.}
\label{tab:qwen-comet}
\end{table}

\begin{table}[!htbp]
\centering
\small
\begin{tabular}{l rr}
\toprule
\textbf{Domain} & \textbf{$\Delta$ Adapt.} & \textbf{$\Delta$ Rating} \\
\midrule
social comments      & \cellcolor{dpos8}+29.2 & \cellcolor{dpos8}+10.0 \\
conversation         & \cellcolor{dpos8}+25.6 & \cellcolor{dpos8}+9.3 \\
social posts         & \cellcolor{dpos8}+13.9 & \cellcolor{dpos4}+4.8 \\
narration            & \cellcolor{dpos8}+10.9 & \cellcolor{dpos4}+5.8 \\
reflection pieces    & \cellcolor{dpos8}+9.3  & \cellcolor{dpos4}+5.3 \\
other misc.          & \cellcolor{dpos4}+7.7  & \cellcolor{dpos4}+4.2 \\
how-to / instr.      & \cellcolor{dpos4}+5.5  & \cellcolor{dpos0}+2.4 \\
web misc.            & \cellcolor{dpos0}+3.9  & \cellcolor{dpos0}+1.6 \\
\bottomrule
\end{tabular}
\caption{Per-domain adaptedness and rating improvement (with $-$ without instructions) for Qwen3.5-27B on Indonesian. The domain ordering matches Gemma: informal domains benefit most.}
\label{tab:qwen-domain}
\end{table}

One notable difference: at the 4B level, Qwen3.5 shows near-zero critical errors on Indonesian (0.2\% without, 0.0\% with instructions), unlike Gemma-3-4B (2.4\% and 4.8\%). This may reflect differences in multilingual training data composition between the two model families, with Qwen having stronger Indonesian coverage at small scales.

\clearpage

\setlength{\textfloatsep}{20pt plus 2pt minus 4pt}
\setlength{\floatsep}{12pt plus 2pt minus 2pt}
\setlength{\intextsep}{12pt plus 2pt minus 2pt}
\renewcommand{\arraystretch}{1.0}

\raggedbottom
\section{Prompts}
\label{app:prompts}

This section lists the prompts used in each stage of the MT+ pipeline (Figure~\ref{fig:pipeline}).
Placeholders are shown in \texttt{\{braces\}}.

\subsection{Instruction Generation (Step a)}
\label{app:prompt-instruction}

For BOUQuET, instructions are generated from metadata (columns \texttt{domain}, \texttt{par\_comment}, \texttt{tags}), and the English source text. For WMT24++, a simplified variant uses domain and source text only.

\begin{promptbox}{System prompt}
You are an expert at creating translation instructions. Given metadata about a text segment (domain, context description, linguistic tags, and a reference translation), generate a concise user instruction that a translator would use to properly translate the source text.

The instruction should:
1.\ Specify the translation setting (e.g., blog post, formal document, social media, novel).
2.\ Define the target audience (e.g., general readers, professionals, fans of a genre).
3.\ Indicate the appropriate tone and style.

Output ONLY the instruction, nothing else. Keep it to 1--3 sentences.
\end{promptbox}

\begin{promptbox}[fontupper=\small\ttfamily]{User message template}
Domain: \{domain\}\\
Context: \{par\_comment\}\\
Linguistic tags: \{tags\}\\
Reference translation: \{eng\_text\}\\[0.3em]
Generate a user instruction for translating this text.
\end{promptbox}

Three fixed few-shot examples are prepended (one each from the web, social posts, and narration domains).

Every instruction is post-edited by paper authors. Edits were minor, with a tendency to make instructions less verbose for clarity.

\subsection{Translation (Step b)}
\label{app:prompt-translation}

We use two prompt variants, with and without a user instruction.

\subsubsection{Without instructions.}
\label{app:prompt-translation-without-instructions}

\begin{promptbox}{System prompt}
You are an expert translator. Translate the English text to \{language\}. Give only the translation, no extra commentary.
\end{promptbox}

\begin{promptbox}[fontupper=\small\ttfamily]{User message}
\{source\}
\end{promptbox}

Both conditions (with/without) include 3 fixed few-shot examples drawn from a held-out split, and matching the format of the user message.

\subsubsection{With instructions.}
\label{app:prompt-translation-with-instructions}

\begin{promptbox}{System prompt}
You are an expert translator. Translate the English text to \{language\}. Follow the user's instruction about tone, style, and context. Give only the translation, no extra commentary.
\end{promptbox}

\begin{promptbox}[fontupper=\small\ttfamily]{User message}
Instruction: \{instruction\}\\[0.3em]
Text: \{eng\_text\}
\end{promptbox}

\subsection{LLM-Judge Evaluation (Step c)}
\label{app:prompt-judge}

\begin{promptbox}{System prompt}
You are an expert translation evaluator. You will be given:
\begin{itemize}[leftmargin=1em,itemsep=0.15em,topsep=0.2em]
\item Instruction: the desired tone/context for the translation
\item Source sentence (English)
\item Translation to evaluate (into \{language\})
\end{itemize}

Your task has three parts:

\textbf{Part 1: Identify Error Spans.}
Identify specific text spans in the translation that contain errors. For each error, provide the exact text and a label: \textbf{Major} (meaning change, mistranslation, wrong language, or omission) or \textbf{Minor} (grammar, style, awkward phrasing, typos). Return an empty list if there are no errors.

\textbf{Part 2: Critical Failure Flag.}
Set to true if the output is not a translation at all (wrong language, unrelated content, or not translating the source). Otherwise, false.

\textbf{Part 3: Rate the Translation} (0--100 scale).

\textit{Translation Rating}: 0 = no meaning preserved; 33 = some meaning, significant errors; 66 = most meaning, minor issues; 100 = perfect.

\textit{Adaptedness}: 0 = not adapted; 33 = minimally; 66 = mostly; 100 = perfectly adapted to the requested tone/context.

Respond ONLY with a valid JSON object:

{\small\ttfamily
\{"error\_spans": [...],\\
\phantom{\ }"is\_critical\_failure": false,\\
\phantom{\ }"translation\_rating\_score": 85,\\
\phantom{\ }"adaptedness\_score": 75,\\
\phantom{\ }"comment": "Brief explanation"\}}
\end{promptbox}

\begin{promptbox}[fontupper=\small\ttfamily]{User message template}
Instruction: \{instruction\}\\
Source sentence (English): \{source\}\\
Translation to evaluate: \{translation\}\\
Evaluate this translation.
\end{promptbox}

\section{Paragraph-Level Translation}
\label{app:paragraph}

To test whether paragraph context alone (without explicit per-sentence instructions) helps translation adaptation, we compare two paragraph prompting strategies against the sentence-level baseline ($\Delta$): $P_1$, which marks a single target sentence for translation within the paragraph, and $P_2$, which translates all sentences at once in a numbered list. Figure~\ref{fig:p1-p2-prompts} illustrates the two strategies on the same English paragraph. We also include the $\Delta$ from adding explicit instructions (Instr) for comparison. Results are shown in Table~\ref{tab:paragraph-comparison}.

\begin{figure*}[!htbp]
\centering
\begin{minipage}[t]{0.49\textwidth}
\centering
\textbf{$P_1$: sentence markers (en$\rightarrow$fra)}\\[0.3em]
\begin{promptbox}[breakable=false,fontupper=\scriptsize\ttfamily]{Prompt}
My flight finally landed.\\
\textbf{[TRANSLATE]}The hotel shuttle was waiting outside.\textbf{[/TRANSLATE]}\\
All I want now is a hot shower.
\end{promptbox}
\vspace{0.2em}
\begin{promptbox}[breakable=false,fontupper=\scriptsize\ttfamily,colback=blue!5,colframe=blue!35]{Model output}
La navette de l'h\^otel attendait dehors.
\end{promptbox}
\end{minipage}\hfill
\begin{minipage}[t]{0.49\textwidth}
\centering
\textbf{$P_2$: numbered list (en$\rightarrow$fra)}\\[0.3em]
\begin{promptbox}[breakable=false,fontupper=\scriptsize\ttfamily]{Prompt}
S1: My flight finally landed.\\
S2: The hotel shuttle was waiting outside.\\
S3: All I want now is a hot shower.
\end{promptbox}
\vspace{0.2em}
\begin{promptbox}[breakable=false,fontupper=\scriptsize\ttfamily,colback=blue!5,colframe=blue!35]{Model output}
S1: Mon vol a enfin atterri.\\
S2: La navette de l'h\^otel attendait dehors.\\
S3: Tout ce que je veux, c'est une douche chaude.
\end{promptbox}
\end{minipage}
\caption{The two paragraph-level prompting strategies illustrated on a 3-sentence paragraph. \textbf{$P_1$}: only the bracketed sentence is returned; the model uses the surrounding sentences purely as context. \textbf{$P_2$}: all sentences are translated together and returned in the same numbered format. Both strategies are scored at the sentence level.}
\label{fig:p1-p2-prompts}
\end{figure*}

\begin{table*}[!htbp]
\centering
\small
\setlength{\tabcolsep}{3pt}
\begin{tabular}{ll rrrrr r rrrrr r}
\toprule
& & \multicolumn{6}{c}{\textbf{Translation Rating}} & \multicolumn{6}{c}{\textbf{Adaptedness}} \\
\cmidrule(lr){3-8} \cmidrule(lr){9-14}
& & \textbf{fra} & \textbf{ind} & \textbf{ukr} & \textbf{khm} & \textbf{jav} & \textbf{Avg.} & \textbf{fra} & \textbf{ind} & \textbf{ukr} & \textbf{khm} & \textbf{jav} & \textbf{Avg.} \\
\midrule
\multirow{4}{*}{Gemma-3-27B}
 & Base & 95.0 & 93.3 & 94.5 & 76.8 & 81.1 & 88.1 & 89.0 & 83.4 & 90.7 & 72.3 & 77.2 & 82.6 \\
 & $\Delta\;P_1-$Base & \cellcolor{dneg0}--2.1 & \cellcolor{dneg0}--2.6 & \cellcolor{dneg4}--3.0 & \cellcolor{dneg4}--4.0 & \cellcolor{dneg4}--3.8 & \cellcolor{dneg4}--3.1 & \cellcolor{dneg0}--0.9 & \cellcolor{dneg0}--1.6 & \cellcolor{dneg0}--1.7 & \cellcolor{dneg4}--4.1 & \cellcolor{dneg4}--3.0 & \cellcolor{dneg0}--2.3 \\
 & $\Delta\;P_2-$Base & \cellcolor{dpos0}+0.6 & \cellcolor{dpos0}+0.2 & \cellcolor{dpos0}+0.6 & \cellcolor{dneg0}--0.9 & \cellcolor{dpos0}+0.5 & \cellcolor{dpos0}+0.2 & \cellcolor{dpos0}+1.7 & \cellcolor{dpos0}+0.4 & \cellcolor{dpos0}+0.4 & \cellcolor{dneg0}--1.0 & \cellcolor{dpos0}+0.1 & \cellcolor{dpos0}+0.3 \\
 & $\Delta$ Instr$-$Base & \cellcolor{dpos0}+2.4 & \cellcolor{dpos4}+4.7 & \cellcolor{dpos0}+1.3 & \cellcolor{dpos0}+1.2 & \cellcolor{dpos0}+0.2 & \cellcolor{dpos0}+2.0 & \cellcolor{dpos8}+9.3 & \cellcolor{dpos8}+15.1 & \cellcolor{dpos4}+6.2 & \cellcolor{dpos4}+7.9 & \cellcolor{dpos4}+7.1 & \cellcolor{dpos8}+9.1 \\
\midrule
\multirow{4}{*}{Qwen3.5-27B}
 & Base & 94.0 & 92.4 & 93.1 & 80.1 & 80.3 & 88.0 & 86.1 & 80.7 & 88.9 & 72.8 & 75.1 & 80.7 \\
 & $\Delta\;P_1-$Base & \cellcolor{dpos0}+0.5 & \cellcolor{dneg0}--0.1 & \cellcolor{dpos0}+0.1 & \cellcolor{dneg0}--0.3 & \cellcolor{dpos0}+0.2 & \cellcolor{dpos0}+0.1 & \cellcolor{dpos0}+1.2 & \cellcolor{dneg0}--0.1 & \cellcolor{dneg0}--0.3 & \cellcolor{dpos0}+0.3 & \cellcolor{dpos0}+0.2 & \cellcolor{dpos0}+0.3 \\
 & $\Delta\;P_2-$Base & \cellcolor{dpos0}+0.9 & \cellcolor{dpos0}+0.1 & \cellcolor{dpos0}+0.7 & \cellcolor{dneg0}--0.2 & \cellcolor{dpos0}+0.6 & \cellcolor{dpos0}+0.4 & \cellcolor{dpos0}+2.1 & \cellcolor{dpos0}+1.2 & \cellcolor{dpos0}+0.0 & \cellcolor{dneg0}--0.3 & \cellcolor{dpos0}+0.8 & \cellcolor{dpos0}+0.8 \\
 & $\Delta$ Instr$-$Base & \cellcolor{dpos4}+4.0 & \cellcolor{dpos4}+5.1 & \cellcolor{dpos0}+1.9 & \cellcolor{dpos0}+2.8 & \cellcolor{dneg0}--2.0 & \cellcolor{dpos0}+2.4 & \cellcolor{dpos8}+11.5 & \cellcolor{dpos8}+16.7 & \cellcolor{dpos4}+7.1 & \cellcolor{dpos8}+9.0 & \cellcolor{dpos4}+4.3 & \cellcolor{dpos8}+9.7 \\
\midrule
\multirow{5}{*}{Gemma-4-31B}
 & Base & 96.0 & 93.7 & 96.4 & 92.1 & 92.4 & 94.1 & 89.1 & 82.8 & 92.9 & 84.2 & 87.0 & 87.2 \\
 & $\Delta\;P_1-$Base & \cellcolor{dpos0}+0.2 & \cellcolor{dpos0}+0.3 & \cellcolor{dpos0}+0.2 & \cellcolor{dpos0}+0.7 & \cellcolor{dpos0}+0.0 & \cellcolor{dpos0}+0.3 & \cellcolor{dpos0}+1.2 & \cellcolor{dpos0}+1.0 & \cellcolor{dpos0}+0.5 & \cellcolor{dpos0}+1.2 & \cellcolor{dpos0}+0.4 & \cellcolor{dpos0}+0.9 \\
 & $\Delta\;P_2-$Base & \cellcolor{dpos0}+0.7 & \cellcolor{dpos0}+0.7 & \cellcolor{dpos0}+0.1 & \cellcolor{dpos0}+1.1 & \cellcolor{dpos0}+0.4 & \cellcolor{dpos0}+0.6 & \cellcolor{dpos0}+2.5 & \cellcolor{dpos0}+2.0 & \cellcolor{dpos0}+0.5 & \cellcolor{dpos0}+1.2 & \cellcolor{dpos0}+1.5 & \cellcolor{dpos0}+1.5 \\
 & $\Delta$ Self-Instr$-$Base & \cellcolor{dpos0}+2.1 & \cellcolor{dpos4}+4.6 & \cellcolor{dpos0}+1.0 & \cellcolor{dpos0}+2.6 & \cellcolor{dpos0}+3.0 & \cellcolor{dpos0}+2.7 & \cellcolor{dpos4}+7.6 & \cellcolor{dpos8}+14.0 & \cellcolor{dpos0}+3.0 & \cellcolor{dpos8}+9.1 & \cellcolor{dpos4}+7.1 & \cellcolor{dpos4}+8.2 \\
 & $\Delta$ Instr$-$Base & \cellcolor{dpos4}+3.2 & \cellcolor{dpos4}+5.3 & \cellcolor{dpos0}+2.2 & \cellcolor{dpos4}+3.4 & \cellcolor{dpos4}+3.7 & \cellcolor{dpos4}+3.6 & \cellcolor{dpos8}+9.9 & \cellcolor{dpos8}+15.9 & \cellcolor{dpos4}+5.8 & \cellcolor{dpos8}+10.6 & \cellcolor{dpos8}+8.8 & \cellcolor{dpos8}+10.2 \\
\midrule
\multirow{4}{*}{GPT-5.4}
 & Base & 96.2 & 93.0 & 97.2 & 91.8 & 89.7 & 93.6 & 88.4 & 82.3 & 92.8 & 82.7 & 80.2 & 85.3 \\
 & $\Delta\;P_1-$Base & \cellcolor{dpos0}+0.2 & \cellcolor{dpos0}+1.1 & \cellcolor{dpos0}+0.8 & \cellcolor{dpos0}+1.3 & \cellcolor{dpos0}+2.3 & \cellcolor{dpos0}+1.1 & \cellcolor{dpos0}+2.2 & \cellcolor{dpos0}+1.4 & \cellcolor{dpos0}+2.1 & \cellcolor{dpos0}+2.3 & \cellcolor{dpos4}+6.3 & \cellcolor{dpos0}+2.9 \\
 & $\Delta\;P_2-$Base & \cellcolor{dpos0}+0.7 & \cellcolor{dpos0}+1.2 & \cellcolor{dpos0}+0.6 & \cellcolor{dpos0}+1.3 & \cellcolor{dpos0}+2.3 & \cellcolor{dpos0}+1.2 & \cellcolor{dpos4}+3.1 & \cellcolor{dpos0}+0.9 & \cellcolor{dpos0}+1.7 & \cellcolor{dpos0}+2.2 & \cellcolor{dpos4}+6.2 & \cellcolor{dpos0}+2.8 \\
 & $\Delta$ Instr$-$Base & \cellcolor{dpos0}+2.6 & \cellcolor{dpos4}+5.8 & \cellcolor{dpos0}+1.8 & \cellcolor{dpos4}+4.0 & \cellcolor{dpos4}+4.3 & \cellcolor{dpos4}+3.7 & \cellcolor{dpos4}+8.2 & \cellcolor{dpos8}+16.0 & \cellcolor{dpos4}+5.3 & \cellcolor{dpos8}+11.1 & \cellcolor{dpos8}+11.2 & \cellcolor{dpos8}+10.4 \\
\bottomrule
\end{tabular}
\caption{Comparison of paragraph prompting strategies ($P_1$ / $P_2$) with explicit instructions about translation purpose (Instr).  Instructions always outperform paragraph-level context, with the gap widening as model capability increases. Base~= sentence-level translation, without instructions. $P_1$~= sentence markers: the target sentence is wrapped in \texttt{[TRANSLATE]} tags and the model translates only that sentence using the rest as context. $P_2$~= numbered list: all sentences are presented as \texttt{S1:~... / S2:~...} and the model returns translations in the same format.}
\label{tab:paragraph-comparison}
\end{table*}

27B models (both Gemma-3 and Qwen3.5) get uneven gains from paragraph-level translation across languages (often negative for Gemma-3 and near zero for Qwen3.5), while explicit instructions yield consistent positive gains across all five languages. Stronger models (Gemma-4-31B and GPT-5.4) extract more from both paragraph context and explicit instructions, with all $\Delta$ values positive.
Across all four models, explicit instructions yield larger adaptedness gains than either paragraph strategy (roughly +9 to +10 points over baseline), confirming that \textbf{explicit purpose instructions remain the most reliable adaptation method across models}.

\section{Instruction Ablation Study}
\label{app:ablation}

We decompose instructions into two components: \textbf{context} (situational framing, e.g. "This is a conversation between two female friends") and \textbf{purpose} (pragmatic directives, e.g. "Use casual, colloquial language") and measure how each contributes to translation adaptedness with Gemma-3-27B.

Across languages we find that \textbf{any instruction condition (context, purpose, or full) improves adaptedness}, with context and purpose only partially closing the adaptedness gap compared to the full instruction (Table~\ref{tab:ablation-gap}).
For higher-resource languages (French, Indonesian, Ukrainian), context alone tends to outperform purpose alone, while for lower-resource languages (Khmer, Javanese), the opposite is true. This points to a model capability, where the model accurately identifies what is the right register or tone to use for a given context for high-resource languages (e.g. "this is a conversation between friends" $\rightarrow$ model uses an informal tone), but needs more explicit guidance for lower-resource languages (e.g. "use casual language" $\rightarrow$ model uses an informal tone).

\begin{figure*}[!htbp]
\centering
\includegraphics[width=\textwidth]{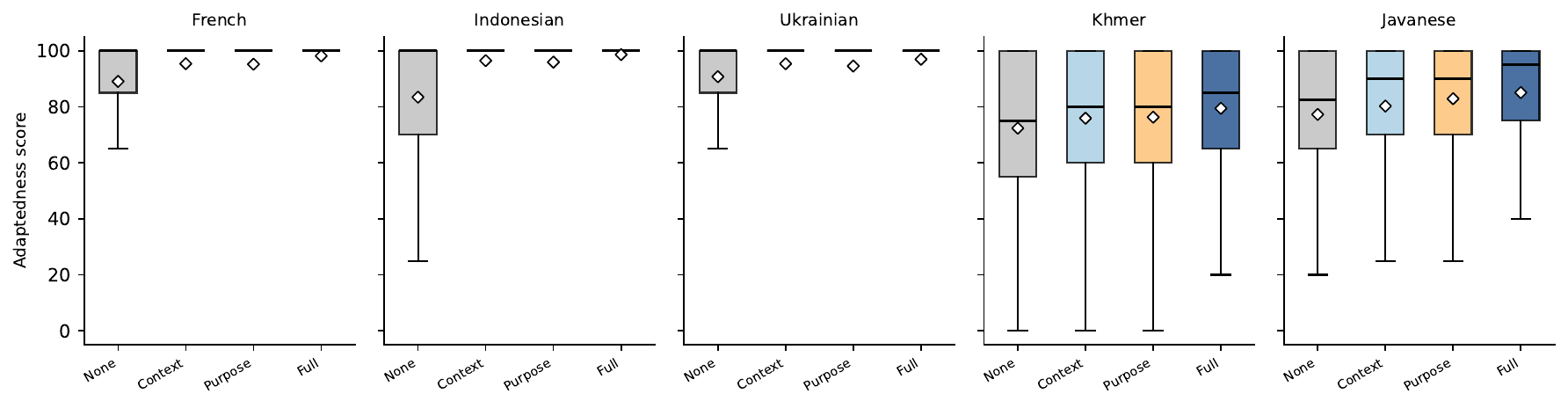}
\caption{Distribution of adaptedness scores by instruction variant (Gemma-3-27B, BOUQuET dev). Diamond markers indicate means. For high-resource languages, any instruction variant collapses the distribution upward; for low-resource languages, variance remains high regardless of variant.}
\label{fig:ablation-boxplot}
\end{figure*}

For all instruction conditions, we find that for higher-resource languages, adaptedness variance is very low when any instruction is added (Figure~\ref{fig:ablation-boxplot}), whereas for \textbf{lower-resource languages, variance is much higher across all conditions}, albeit lower for the full instruction variant.
This suggests that for lower-resource languages, the model's pragmatic competence is less robust: even explicit instructions cannot reliably steer register and tone.

\begin{table*}[!htbp]
\centering
\begin{minipage}[t]{0.48\textwidth}
\centering
\small
\begin{tabular}{l rrrr}
\toprule
\textbf{Lang.} & \textbf{None} & \textbf{Context} & \textbf{Purpose} & \textbf{Full} \\
\midrule
fra     & 89.0\,\scriptsize{$\pm$19.1} & 95.4\,\scriptsize{$\pm$14.1} & 95.2\,\scriptsize{$\pm$13.0} & 98.3\,\scriptsize{$\pm$\phantom{0}8.0} \\
ind & 83.4\,\scriptsize{$\pm$22.7} & 96.5\,\scriptsize{$\pm$10.8} & 95.9\,\scriptsize{$\pm$11.0} & 98.6\,\scriptsize{$\pm$\phantom{0}5.9} \\
ukr  & 90.7\,\scriptsize{$\pm$17.0} & 95.4\,\scriptsize{$\pm$12.6} & 94.5\,\scriptsize{$\pm$14.1} & 96.9\,\scriptsize{$\pm$10.2} \\
khm      & 72.3\,\scriptsize{$\pm$24.0} & 75.9\,\scriptsize{$\pm$23.6} & 76.3\,\scriptsize{$\pm$23.4} & 80.2\,\scriptsize{$\pm$22.1} \\
jav   & 77.2\,\scriptsize{$\pm$22.9} & 80.2\,\scriptsize{$\pm$23.0} & 82.9\,\scriptsize{$\pm$21.5} & 84.4\,\scriptsize{$\pm$20.4} \\
\bottomrule
\end{tabular}
\captionof{table}{Mean adaptedness score ($\pm$ std.\ dev.) by instruction variant (Gemma-3-27B, BOUQuET dev). High-resource languages show higher means and reduced variance with any instruction; low-resource languages see modest mean gains with persistent variance.}
\label{tab:ablation-stddev}
\end{minipage}%
\hfill
\begin{minipage}[t]{0.48\textwidth}
\centering
\small
\begin{tabular}{l cc}
\toprule
\textbf{Language} & \textbf{Context} & \textbf{Purpose} \\
\midrule
French     & \cellcolor{win65} 69\% \scriptsize{(+6.4)}  & \cellcolor{win65} 67\% \scriptsize{(+6.2)} \\
Indonesian & \cellcolor{win75} 86\% \scriptsize{(+13.1)} & \cellcolor{win75} 82\% \scriptsize{(+12.5)} \\
Ukrainian  & \cellcolor{win75} 76\% \scriptsize{(+4.7)}  & \cellcolor{win65} 61\% \scriptsize{(+3.8)} \\
Khmer      & \cellcolor{win45} 46\% \scriptsize{(+3.6)}  & \cellcolor{win55} 51\% \scriptsize{(+4.0)} \\
Javanese   & \cellcolor{win45} 42\% \scriptsize{(+3.0)}  & \cellcolor{win75} 79\% \scriptsize{(+5.7)} \\
\bottomrule
\end{tabular}
\captionof{table}{Percentage of the adaptedness gap closed toward full instructions, with absolute score improvement over no-instruction baseline in parentheses. For high-resource languages, context alone recovers 69--86\% of the full benefit. For Javanese, purpose only wins largely over context (79\% vs.\ 42\%), while Khmer and Ukrainian show balanced contributions from both.}
\label{tab:ablation-gap}
\end{minipage}
\end{table*}

\section{Self-Instruction: Per-language \& WMT24++ Results}
\label{app:self-instr-gemma3}

In this section, we report the per-language performance and WMT24++ replication of the self-instruction experiment from \S\ref{sec:self-instr}.

For \textbf{Gemma-4-31B} (Table~\ref{tab:self-instr-gemma4}), Self-Instr is positive for every language on both metrics, with per-language gap closures strongest on Indonesian (88\%/87\% on adapt./rating) and weakest on Ukrainian (52\%/45\%).

For \textbf{Gemma-3-27B} (Table~\ref{tab:self-instr-gemma3}), Self-Instr closes 71\% of the adaptedness gap and 43\% of the translation-rating gap on average: weaker than Gemma-4-31B but still positive for every language on adaptedness, confirming that the effect is not tied to Gemma-4. The translation-rating gap closes less consistently, with a regression on Javanese (Self-Instr~79.5 vs.\ None~81.1 while Full reaches~81.3), which we attribute to Gemma-3-27B's weaker Javanese performance.

\begin{table*}[!t]
\centering
\small
\begin{tabular}{l rrrr rrrr}
\toprule
 & \multicolumn{4}{c}{\textbf{Adaptedness}} & \multicolumn{4}{c}{\textbf{Translation rating}} \\
\cmidrule(lr){2-5} \cmidrule(lr){6-9}
\textbf{Language} & \textbf{None} & \textbf{Self-Instr} & \textbf{Full} & \textbf{\% Gap} & \textbf{None} & \textbf{Self-Instr} & \textbf{Full} & \textbf{\% Gap} \\
\midrule
French     & 89.1 & 96.7 & 99.0 & \cellcolor{dpos8}77\% & 96.0 & 98.1 & 99.2 & \cellcolor{dpos4}66\% \\
Indonesian & 82.8 & 96.8 & 98.8 & \cellcolor{dpos8}88\% & 93.7 & 98.3 & 99.0 & \cellcolor{dpos8}87\% \\
Ukrainian  & 92.9 & 95.9 & 98.7 & \cellcolor{dpos4}52\% & 96.4 & 97.4 & 98.6 & \cellcolor{dpos4}45\% \\
Khmer      & 84.2 & 93.3 & 94.8 & \cellcolor{dpos8}86\% & 92.1 & 94.7 & 95.5 & \cellcolor{dpos8}76\% \\
Javanese   & 87.0 & 94.1 & 95.9 & \cellcolor{dpos8}80\% & 92.4 & 95.4 & 96.0 & \cellcolor{dpos8}83\% \\
\bottomrule
\end{tabular}
\caption{Per-language self-instruction results with \textbf{Gemma-4-31B} on BOUQuET dev, under \textbf{None} (no instruction), \textbf{Self-Instr} (self-generated from paragraph), and \textbf{Full} (gold). \textbf{\% Gap} is the fraction of the None-to-Full gap closed by Self-Instr.}
\label{tab:self-instr-gemma4}
\end{table*}

\begin{table*}[!t]
\centering
\small
\begin{tabular}{l rrrr rrrr}
\toprule
 & \multicolumn{4}{c}{\textbf{Adaptedness}} & \multicolumn{4}{c}{\textbf{Translation rating}} \\
\cmidrule(lr){2-5} \cmidrule(lr){6-9}
\textbf{Language} & \textbf{None} & \textbf{Self-Instr} & \textbf{Full} & \textbf{\% Gap} & \textbf{None} & \textbf{Self-Instr} & \textbf{Full} & \textbf{\% Gap} \\
\midrule
French     & 89.0 & 95.7 & 98.3 & \cellcolor{dpos8}72\% & 95.0 & 96.6 & 97.4 & \cellcolor{dpos4}67\% \\
Indonesian & 83.4 & 95.7 & 98.6 & \cellcolor{dpos8}81\% & 93.3 & 96.8 & 98.0 & \cellcolor{dpos8}74\% \\
Ukrainian  & 90.7 & 94.6 & 96.9 & \cellcolor{dpos4}63\% & 94.5 & 95.0 & 95.7 & \cellcolor{dpos0}42\%$^\dagger$ \\
Khmer      & 72.3 & 77.6 & 80.2 & \cellcolor{dpos4}67\% & 76.8 & 77.0 & 78.0 & \cellcolor{dpos0}17\%$^\dagger$ \\
Javanese   & 77.2 & 81.4 & 84.4 & \cellcolor{dpos4}58\% & 81.1 & 79.5 & 81.3 & \cellcolor{dneg4}regress.$^\ddagger$ \\
\bottomrule
\end{tabular}
\caption{Per-language self-instruction results with \textbf{Gemma-3-27B} on BOUQuET dev. Same metrics and setup as Table~\ref{tab:self-instr-gemma4}, with Gemma-3-27B used both to generate the self-instruction and to translate. $^\dagger$Full improves translation rating by less than 1.5 points over None, making the ratio noisy. $^\ddagger$Self-Instr mean (79.5) falls below None (81.1), while Full reaches 81.3.}
\label{tab:self-instr-gemma3}
\end{table*}

\paragraph{Replication on WMT24++}
For each segment, Gemma-3-27B first drafts a user instruction from surrounding document (target sentence marked with \texttt{[TARGET]}), then translates conditioned on its own brief; as in \S\ref{sec:self-instr}, adaptedness is judged against the curated instruction so rating is comparable across conditions. Table~\ref{tab:self-instr-wmt-27b} shows the BOUQuET finding replicates closely: Self-Instr closes 74\% of the adaptedness gap and 36\% of the translation-rating gap on average, against 71\%/43\% for the same model on BOUQuET, and is positive for every language on both metrics. The effect is notable given that raw document context, fed directly as $P_1$ paragraph input, \textit{hurt} adaptedness by --10.6 on these same languages (Table~\ref{tab:full-wmt-27b}): the same document context flips from harmful to beneficial once the model externalises it as an explicit brief. WMT24++ documents are longer than BOUQuET paragraphs, yet the model still drafts useful briefs from them. Curated instructions retain an edge (a stronger drafting model and an explicit domain label), but a \textbf{model with no user-supplied brief recovers most of the adaptedness benefit from document context alone}.

\begin{table*}[!t]
\centering
\small
\begin{tabular}{l rrrr rrrr}
\toprule
 & \multicolumn{4}{c}{\textbf{Adaptedness}} & \multicolumn{4}{c}{\textbf{Translation rating}} \\
\cmidrule(lr){2-5} \cmidrule(lr){6-9}
\textbf{Language} & \textbf{None} & \textbf{Self-Instr} & \textbf{Full} & \textbf{\% Gap} & \textbf{None} & \textbf{Self-Instr} & \textbf{Full} & \textbf{\% Gap} \\
\midrule
French & 83.0 & 92.0 & 96.9 & \cellcolor{dpos4}64\% & 89.1 & 90.7 & 96.5 & \cellcolor{dpos0}22\% \\
Indonesian & 78.4 & 93.5 & 96.9 & \cellcolor{dpos8}82\% & 86.8 & 91.8 & 96.2 & \cellcolor{dpos4}53\% \\
Ukrainian & 81.6 & 91.4 & 95.2 & \cellcolor{dpos8}72\% & 87.4 & 89.3 & 94.1 & \cellcolor{dpos0}28\% \\
\bottomrule
\end{tabular}
\caption{Per-language self-instruction results with Gemma-3-27B on \textbf{WMT24++} (fra/ind/ukr, 997~items per language), under \textbf{None} (no instruction), \textbf{Self-Instr} (self-generated from the surrounding document), and \textbf{Full} (curated). \textbf{\% Gap} is the fraction of the None-to-Full gap closed by Self-Instr. Translations were produced with the self-instruction; adaptedness is judged against the curated instruction. None and Full match the Baseline and Instructions rows of Table~\ref{tab:full-wmt-27b}.}
\label{tab:self-instr-wmt-27b}
\end{table*}

\section{PEFT with a 27B teacher}
\label{app:lora-sft}

\paragraph{Recipe.} To mitigate the high rate of critical failure rates for the 4B model on lower-resource languages, we experiment with knowledge distillation from the 27B model, using a LoRA adapter.

\begin{figure*}[!htbp]
\centering
\begin{tikzpicture}[
    box/.style={draw, rounded corners, fill=#1!12, minimum height=0.7cm, minimum width=1.9cm, font=\footnotesize, align=center, inner sep=3pt},
    box/.default=blue,
    llm/.style={draw, circle, fill=orange!15, minimum size=1cm, font=\footnotesize, align=center, inner sep=1pt},
    arr/.style={-{Stealth[length=4pt]}, semithick},
    stagelbl/.style={font=\footnotesize\itshape, anchor=south},
]
\node[box=blue] (src) at (0, 0) {SMOL\\EN sentence};

\node[stagelbl] at (3.1, 1.55) {Draft instruction};
\node[llm] (genllm) at (3.1, 0.7) {Gemini};
\node[box=orange] (instr) at (6.0, 0.7) {Instruction};

\node[stagelbl] at (9.3, 1.55) {Teacher translate};
\node[llm] (teacher) at (9.3, 0) {\small Gemma-3\\27B};
\node[box=green!70!black] (tgt) at (12.2, 0) {Teacher\\translation};

\node[stagelbl] at (9.3, -1.35) {SFT (LoRA)};
\node[llm] (student) at (9.3, -2.2) {\small Gemma-3\\4B};
\node[box=red!70!black] (adapter) at (12.2, -2.2) {LoRA\\adapter};

\draw[arr] (src.east) -| (genllm.south);
\draw[arr] (genllm.east) -- (instr.west);
\draw[arr] (instr.east) -| (teacher.north west);
\draw[arr] (src.east) -- (teacher.west);
\draw[arr] (teacher.east) -- (tgt.west);

\draw[arr] (tgt.south) -- (adapter.north);
\draw[arr] (src.south) |- (student.west);
\draw[arr] (instr.south) |- (student.north west);
\draw[arr] (student.east) -- (adapter.west);
\end{tikzpicture}
\caption{PEFT pipeline with a 27B teacher. For each English sentence from \texttt{google/smol}, Gemini drafts a register/tone instruction; Gemma-3-27B translates conditioned on the sentence and the instruction. The (source, instruction, teacher-translation) triple is used for supervised SFT of a LoRA adapter for Gemma-3-4B, with assistant-only loss on the teacher output.}
\label{fig:lora-sft-pipeline}
\end{figure*}
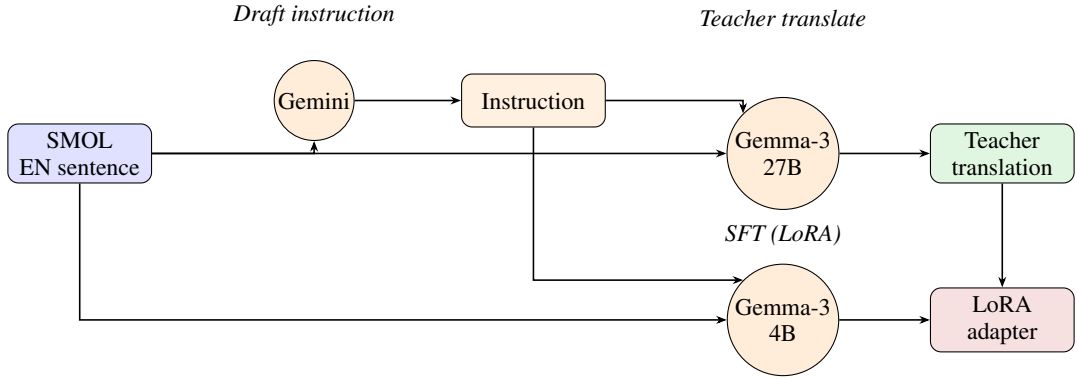

\begin{enumerate}[itemsep=0.25em]
    \item \textbf{Sources.} We use all English source sentences from SMOL \cite{caswell-etal-2025-smol}\footnote{\href{https://huggingface.co/datasets/google/smol}{hf.co/datasets/google/smol}}, a translation training dataset that balances data diversity and quality. We use both SmolDoc (7,805 sentences) and SmolSent (863 sentences), which have 7,815 unique English source sentences after deduplication.

    \item \textbf{Instructions.} For each source sentence, Gemini drafts a concise, language-agnostic instruction, similar to that in BOUQuET (\S\ref{sec:data}).
    \item \textbf{Teacher.} Gemma-3-27B translates each (source, instruction) pair with the same prompt as our main experiments (\S\ref{app:prompt-translation}).
    \item \textbf{Student.} Gemma-3-4B with a LoRA adapter (r=16, $\alpha$=32, dropout~0.05), is trained for 1~epoch at learning rate $2 \times 10^{-5}$ with batch size~4 and gradient accumulation~16. We use the conversational chat format (\textit{user}: instruction + source; \textit{assistant}: teacher translation) and mask all non-assistant tokens (assistant-only loss).
\end{enumerate}
We train one adapter for Ukrainian and one for Javanese, being the two languages with high critical failure rates for the 4B model with instructions (16.5\% and 45.5\%, respectively).

\begin{table*}[!htbp]
\centering
\small
\begin{tabular}{l l l rrr}
\toprule
\textbf{Lang} & \textbf{Model} & \textbf{Instr.} & \textbf{Adapt.} & \textbf{Rating} & \textbf{Crt.\%} \\
\midrule
\multirow{5}{*}{ukr}
 & 4B base         & w/o & 77.5 & 81.3 & 2.6 \\
 & 4B base         & w/ & 74.6 & 71.6 & 16.5 \\
 & 4B + SFT-ukr    & w/o & \textbf{82.9} \goodup{5.4} & \textbf{85.5} \goodup{4.2} & \textbf{0.6} \gooddown{-2.0} \\
 & 4B + SFT-ukr    & w/ & \textbf{88.8} \verygoodup{14.2} & \textbf{87.4} \verygoodup{15.8} & \textbf{0.0} \gooddown{-16.5} \\
 & 4B + SFT-jav    & w/ & 83.8 \goodup{9.2} & 82.6 \verygoodup{11.0} & 1.6 \gooddown{-14.9} \\
\midrule
\multirow{5}{*}{jav}
 & 4B base         & w/o & 39.3 & 43.3 & 7.0 \\
 & 4B base         & w/ & 29.0 & 25.2 & 45.5 \\
 & 4B + SFT-jav    & w/o & \textbf{46.6} \goodup{7.3} & \textbf{47.3} \goodup{4.0} & \textbf{4.4} \gooddown{-2.6} \\
 & 4B + SFT-jav    & w/ & \textbf{50.4} \verygoodup{21.4} & \textbf{47.8} \verygoodup{22.6} & \textbf{4.0} \gooddown{-41.5} \\
 & 4B + SFT-ukr    & w/ & 45.1 \verygoodup{16.1} & 43.3 \verygoodup{18.1} & 7.6 \gooddown{-37.9} \\
\bottomrule
\end{tabular}
\caption{PEFT results on BOUQuET dev. SFT vastly reduces critical error rates when instructions are present for both languages (ukr 16.5\% $\rightarrow$ 0.0\%, jav 45.5\% $\rightarrow$ 4.0\%), with cross-lingual transfer (ukr adapter works for jav, and vice-versa). Deltas in each SFT row are against the 4B base in the same instruction condition.}
\label{tab:lora-sft}
\end{table*}

\paragraph{Results.} Three findings (Table~\ref{tab:lora-sft}):
\begin{itemize}
    \item \textbf{Critical failures on instruction-driven translation are almost eliminated.} Ukrainian drops from 16.5\%~$\rightarrow$~0.0\%; Javanese from 45.5\%~$\rightarrow$~4.0\%. Rating and adaptedness both rise substantially (Ukrainian: +15.8 rating, +14.2 adaptedness; Javanese: +22.6 rating, +21.4 adaptedness).
    \item \textbf{Cross-language transfer.} Applying the Ukrainian-trained adapter to Javanese (\textit{4B + SFT-ukr} in Table~\ref{tab:lora-sft}), the 4B model stays in Javanese rather than falling back to Ukrainian, and recovers most of the in-language adapter's benefit: 45.5\%~$\rightarrow$~7.6\% critical (vs.~4.0\% for the Javanese-trained adapter), with +18.1 rating and +16.1 adaptedness over the instructed base. This suggests \textbf{the adapter primarily adds a language-agnostic instruction-following capability} (follow the instruction brief, do not fall back to English), rather than target-language-specific translation knowledge.
    \item \textbf{No forgetting of the without-instruction condition.} Although every training example includes an instruction, the adapter improves even the without-instruction BOUQuET condition (Ukrainian: 2.6\%~$\rightarrow$~0.6\% critical, +4.2 rating; Javanese: 7.0\%~$\rightarrow$~4.4\%, +4.0 rating).
\end{itemize}

Overall, our results suggest that the adapter primarily adds a language-agnostic instruction-following capability, enabling the 4B model to translate both with and without instructions more reliably.

\section{Judge-Input Ablation}
\label{app:judge-inputs}

Our judge sees the source, the instruction, and the translation, and scores both criteria from that single view (Figure~\ref{fig:pipeline}(c)). Adaptedness necessarily requires the instruction, but translation rating may not depend on it. We therefore re-judge with each criterion's input restricted: \textit{source-only} hides the instruction and scores translation rating alone, and \textit{instruction-only} hides the source and scores adaptedness alone. Both run on BOUQuET with Gemma-3-27B and the same Gemini-3-Flash judge, for French and Khmer, the highest and lowest agreement languages in Table~\ref{tab:human-llm-agreement}.

Table~\ref{tab:judge-inputs} shows the two criteria behave differently. Adaptedness is robust: hiding the source leaves the gain from instructions intact in both languages (+12.7 and +7.2, against +9.3 and +7.9 jointly). Translation rating is not: under an instruction-blind judge the small gain we report reverses, to $-2.3$ on French and $-3.4$ on Khmer. Instructions do change the surface form of a translation, and a judge that cannot see why tends to read those changes as departures from the source.

Two caveats bound how far this should be interpreted. First, the source-only condition gives the judge strictly less context than our main setup: it sees one sentence at a time, with no instruction to indicate the intended audience or purpose, so some of the drop may reflect an under-informed judge rather than a worse translation. Second, our human validation (Table~\ref{tab:human-llm-agreement}) was collected for the joint setup only. We have no human annotations judged source-only, so we cannot say whether the instruction-blind judge tracks human ratings as well as the joint judge does. We therefore read this as a bound on the translation-rating result rather than a replacement for it: the adaptedness finding, which is our main claim, survives the separation, while the smaller translation-rating gain depends on the judge seeing the instruction.

\begin{table}[t]
\centering
\small
\setlength{\tabcolsep}{4pt}
\begin{adjustbox}{max width=\columnwidth}
\begin{tabular}{ll rrr rrr}
\toprule
& & \multicolumn{3}{c}{\textbf{fra}} & \multicolumn{3}{c}{\textbf{khm}} \\
\cmidrule(lr){3-5} \cmidrule(lr){6-8}
\textbf{Judge sees} & \textbf{Criterion} & w/o & w/ & $\Delta$ & w/o & w/ & $\Delta$ \\
\midrule
Source + instr.\ & Rating & 95.0 & 97.4 & \cellcolor{dpos4}+2.4 & 76.8 & 78.0 & \cellcolor{dpos0}+1.2 \\
Source only      & Rating & 98.7 & 96.3 & \cellcolor{dneg4}--2.3 & 84.8 & 81.4 & \cellcolor{dneg4}--3.4 \\
\midrule
Source + instr.\ & Adapt. & 89.0 & 98.3 & \cellcolor{dpos8}+9.3 & 72.3 & 80.2 & \cellcolor{dpos8}+7.9 \\
Instruction only & Adapt. & 80.4 & 93.1 & \cellcolor{dpos8}+12.7 & 70.2 & 77.4 & \cellcolor{dpos8}+7.2 \\
\bottomrule
\end{tabular}
\end{adjustbox}
\caption{Judge-input ablation, Gemma-3-27B on BOUQuET (504 items per cell). ``w/o'' and ``w/'' are scores without and with user instructions at translation time. Adaptedness survives hiding the source; translation rating reverses when the instruction is hidden. The source-only judge was not validated against human annotations; see the caveats in \S\ref{app:judge-inputs}.}
\label{tab:judge-inputs}
\end{table}

\setlength{\textfloatsep}{8pt plus 2pt minus 2pt}
\setlength{\floatsep}{6pt plus 2pt minus 2pt}
\setlength{\intextsep}{6pt plus 2pt minus 2pt}
\setlength{\dbltextfloatsep}{8pt plus 2pt minus 2pt}
\setlength{\dblfloatsep}{6pt plus 2pt minus 2pt}
\renewcommand{\arraystretch}{0.9}

\section{Controlled-MT Benchmarks}
\label{app:controlled-mt}

Our instructions are free text, whereas the controllable-MT literature (\S\ref{sec:related}) steers translation on a specific attribute (formality, gender, etc). This section asks whether our findings survive on that literature's own benchmarks and metrics, where scoring is reference-based or uses gold phrase-level annotations, and no LLM judge is involved. We evaluate Gemma-3-27B on CoCoA-MT \citep{nadejde-etal-2022-cocoa} for formality and MT-GenEval \citep{currey-etal-2022-mt} for gender.

\begin{table*}[t]
\centering
\small
\begin{tabular}{ll rrrrrrr rr}
\toprule
& & \multicolumn{7}{c}{\textbf{C0 formal default (\%)}} & & \\
\cmidrule(lr){3-9}
\textbf{CoCoA domain} & \textbf{WMT24++ domain} & de & es & fr & hi & it & ja & nl & \textbf{$\rho$} & \textbf{$p$} \\
\midrule
topical\_chat & social \hfill (matched)  & 10.6 & 2.7 & 65.8 & 68.2 & 1.4 & 36.8 & 0.0 & $-0.964$ & 0.003 \\
telephony     & speech \hfill (matched)  & 9.9 & 6.2 & 59.7 & 51.8 & 3.9 & 26.7 & 4.3 & $-0.821$ & 0.034 \\
call\_center  & news \hfill (placebo)    & 78.0 & 56.3 & 95.3 & 95.0 & 29.4 & 69.9 & 29.4 & $-0.071$ & 0.906 \\
\bottomrule
\end{tabular}
\caption{Does a model's default register predict its un-instructed adaptedness? Cells give the C0 formal-default rate: the share of \textit{un-instructed} outputs rendered formally, among the segments whose output carries one of CoCoA's annotated formality markers. $\rho$ is the Spearman correlation, across the seven languages of a row, between that rate and the WMT24++ baseline adaptedness of the matched domain, with $p$ from an exact permutation test. Negative $\rho$ means a more formal default goes with worse un-instructed adaptedness. News is a placebo: a formal default is appropriate there.}
\label{tab:mechanism}
\end{table*}

\paragraph{Conditions.}
On CoCoA-MT we compare four ways of requesting a register: no control (C0), the categorical tag used by prior work (``Use formal/informal register.'', C1), a BOUQuET-style brief that names the register alongside audience and domain (C2), and an \textit{implicit scenario} that describes the social situation in free text without ever naming a register (C3). C1--C3 each run twice, once targeting each register; C0 requests no register and is reported in Table~\ref{tab:mechanism}. On MT-GenEval we use the counterfactual subset (gender cue inside the sentence; plain translation vs.\ a brief stating the referent's gender) and the contextual subset, where the cue appears \textit{only} in the preceding sentence. The contextual subset mirrors our main conditions directly: sentence alone, paragraph context ($P_1$), explicit instruction, and self-instruction (as in Figure~\ref{fig:self-instr-pipeline}).

\begin{table}[!htbp]
\centering
\small
\setlength{\tabcolsep}{3pt}
\begin{adjustbox}{max width=\columnwidth}
\begin{tabular}{l rrrrrrrr}
\toprule
\textbf{Condition} & de & es & fr & hi & it & ja & nl & pt \\
\midrule
C1 categorical tag & 99.0 & 97.4 & 93.6 & 89.2 & 98.2 & 87.0 & 98.2 & 97.6 \\
C2 explicit brief & 93.5 & 97.0 & 78.9 & 71.1 & 97.7 & 79.4 & 98.5 & 91.2 \\
C3 implicit scenario & 84.6 & 76.1 & 89.8 & 91.3 & 77.3 & 82.0 & 77.6 & 86.0 \\
C3b implicit scenario\smark & 97.7 & 90.9 & 97.4 & 92.9 & 95.4 & 84.8 & 88.7 & 96.9 \\
\midrule
Coverage (\%) & 72 & 73 & 70 & 78 & 59 & 39 & 81 & 69 \\
\bottomrule
\end{tabular}
\end{adjustbox}
\caption{CoCoA-MT formality accuracy, Gemma-3-27B. Mean of M-Acc under a formal and an informal target (50 = chance). C3 and C3b never name the register; they describe the social situation only. Coverage is the share of segments that M-Acc can score, i.e.\ those whose output contains one of CoCoA's annotated formality markers; on the rest the output commits to neither register and is dropped from the M-Acc denominator (reported for C3). \smark~C3b is C3 with a single prompt changed. C3's formal \textit{topical\_chat} scenario, a public discussion forum between strangers, does not imply formality in most of these languages, so C3b substitutes a professional-conference setting; every other scenario, including all three informal ones, is unchanged; see \S\ref{par:lang-specific-norms}.}
\label{tab:cocoa-control}
\end{table}

\paragraph{Explicit control is near ceiling (95.0); implicit control nearly matches it (93.1).}
Table~\ref{tab:cocoa-control} reports control accuracy, the mean of M-Acc under a formal and an informal target, where 50 is chance. Averaged over the eight languages, the categorical tag reaches 95.0 and the implicit scenario 93.1 (C3b), so a description of the social situation that never names the register recovers 96\% of the tag's above-chance control. This is the capability that free-text instructions add over categorical labels: the user describes the situation, and the model chooses the register.

\paragraph{Implicit scenarios expose language-specific norms.}
\label{par:lang-specific-norms}
Our first attempt at a \textit{topical\_chat} scenario, ``a public discussion forum between two people who have just met'', failed in most languages: it reaches 100.0\% formal in German under the explicit tag but only 7.4\% under the scenario, and 2.1\% in Spanish, 2.2\% in Italian. It did not fail everywhere, though: Hindi reaches 89.8\% and Japanese 72.3\%. Hindi and Japanese grammaticalise deference to strangers, whereas German, Spanish, Italian address strangers online informally. The scenario  therefore has no single cross-lingual gold label, exposing a limit in this benchmark, and showing an ability for models to integrate these norms. C3b replaces that cell with a professional-conference scenario, which is more universally formal, and recovers 77.5--100.0\% across languages. C3b was authored after observing C3, and we report both. The failure pattern is informative: \textbf{the model and applies language-specific norms on formality in online forums}, which categorical tags would miss.

\begin{table}[!htbp]
\centering
\small
\setlength{\tabcolsep}{3pt}
\begin{adjustbox}{max width=\columnwidth}
\begin{tabular}{ll rrrrrrrrr}
\toprule
\textbf{Condition} & \textbf{Metric} & ar & de & es & fr & hi & it & pt & ru & \textbf{Avg.} \\
\midrule
Baseline & Strict acc. & 68.7 & 64.0 & 55.3 & 67.7 & 36.0 & 48.3 & 65.7 & 68.0 & 59.2 \\
 & Official acc. & 83.7 & 71.3 & 71.3 & 70.7 & 68.3 & 65.7 & 70.3 & 80.7 & 72.8 \\
\midrule
\textbf{Gender brief} & Strict acc. & 70.3 & 71.7 & 63.7 & 75.0 & 40.0 & 56.3 & 72.3 & 72.3 & \textbf{65.2} \\
 & Official acc. & 88.3 & 78.7 & 80.3 & 79.3 & 72.3 & 73.0 & 76.3 & 86.3 & \textbf{79.3} \\
\bottomrule
\end{tabular}
\end{adjustbox}
\caption{MT-GenEval counterfactual subset, where the gender cue is already in the sentence. Official accuracy is the metric of \citet{currey-etal-2022-mt}, which marks any hypothesis without a wrong-gender word as correct. Strict accuracy additionally requires a correct-gender word.}
\label{tab:geneval-counterfactual}
\end{table}

\paragraph{Instructions beat paragraph context.}
On the MT-GenEval contextual subset (Table~\ref{tab:geneval-context}) the gender cue exists only in the previous sentence, so this dataset is designed to make document context count. While paragraph context lifts strict gender accuracy from 47.0 to 67.5, explicit instructions do better, at 72.9. This replicates \S\ref{sec:fewshot} against gold annotations with no LLM judge in the loop. On the counterfactual subset (Table~\ref{tab:geneval-counterfactual}), where the cue is already inside the sentence, a brief that restates the referent's gender still lifts strict accuracy from 59.2 to 65.2 on average, and the official metric from 72.8 to 79.3.

\begin{table}[!htbp]
\centering
\small
\setlength{\tabcolsep}{3pt}
\begin{adjustbox}{max width=\columnwidth}
\begin{tabular}{l rrrrrrrrr}
\toprule
\textbf{Condition} & ar & de & es & fr & hi & it & pt & ru & \textbf{Avg.} \\
\midrule
Sentence only & 45.3 & 47.1 & 51.5 & 49.9 & 42.9 & 47.9 & 50.2 & 41.1 & 47.0 \\
Paragraph context ($P_1$) & 66.6 & 67.3 & 76.6 & 73.0 & 56.8 & 65.6 & 73.7 & 60.2 & 67.5 \\
\textbf{Instruction} & 72.2 & 72.5 & 81.9 & 77.9 & 65.0 & 72.2 & 77.0 & 64.8 & \textbf{72.9} \\
Self-instruction & 58.7 & 60.7 & 70.6 & 64.2 & 53.3 & 58.4 & 66.9 & 53.1 & 60.7 \\
\bottomrule
\end{tabular}
\end{adjustbox}
\caption{MT-GenEval contextual subset, strict gender accuracy. The gender cue appears only in the preceding sentence, so the sentence-only condition must guess. Strict accuracy requires the hypothesis to commit to the correct gender (the official metric scores a non-translation as correct).}
\label{tab:geneval-context}
\end{table}

\paragraph{Metric guards.}
Both official metrics can be gamed by not committing. MT-GenEval's accuracy marks a hypothesis correct whenever it contains no wrong-gender word, so an empty output scores 1.000 and passing the English source through scores 0.907--1.000 across the eight languages. We therefore report \textit{strict} accuracy, which additionally requires a correct-gender word, alongside the official metric. Both accuracies also err the other way: a hypothesis is flagged whenever it contains any word found only in the opposite-gender reference, articles included, so \textit{le 17 mars} is flagged as masculine where the reference wrote \textit{du 17 mars}. Absolute accuracies are therefore conservative; the comparison across conditions is the meaningful quantity. Similarly, CoCoA-MT's M-Acc drops segments where no marker matches, so a model that dodges hard segments scores well on a small denominator. These issues underline the benefits of using an LLM-judge, similar to our main experiments.

\paragraph{On LLMs defaulting to formal register.}
\S\ref{sec:discussion} argues that instructions help most on conversational text because LLMs default to the written register of their pretraining distribution. CoCoA's C0 condition measures that default directly, with gold annotations, and conditions on segments that require a register choice. The default is strongly domain-conditioned: German is 78.0\% formal on \textit{call\_center} but 10.6\% on \textit{topical\_chat}, so we measure it per language and domain (Table~\ref{tab:mechanism}). We can therefore test the mechanism, by correlating each language's formal-default rate with its \textit{un-instructed} WMT24++ adaptedness, within domain.\footnote{Outcome values come from the batch judge run over WMT24++; seven of the eight languages have no other run. Portuguese is excluded because its WMT24++ outputs were produced with a region-stripped prompt, so its predictor and outcome would describe different language variants.} The two matched pairs are strongly negative and the placebo is null: \textbf{a formal default predicts poor un-instructed adaptedness on conversational text}, and does not on news, where a formal default is correct.

\setlength{\textfloatsep}{20pt plus 2pt minus 4pt}
\setlength{\floatsep}{12pt plus 2pt minus 2pt}
\setlength{\intextsep}{12pt plus 2pt minus 2pt}
\setlength{\dbltextfloatsep}{20pt plus 2pt minus 4pt}
\setlength{\dblfloatsep}{12pt plus 2pt minus 2pt}
\renewcommand{\arraystretch}{1.0}

\section{Annotation Protocol}
\label{app:annotation-protocol}

This appendix details the human-annotation procedure underlying the LLM--human agreement results in Table~\ref{tab:human-llm-agreement} and the broader discussion in \S\ref{sec:eval}.

\paragraph{Annotator pool.}
One native-speaker annotator per target language, recruited through the authors' personal networks:
\begin{itemize}[itemsep=0.15em, topsep=0.2em]
    \item French (French native speaker),
    \item Indonesian (Indonesian native speaker),
    \item Ukrainian (Ukrainian native speaker),
    \item Khmer (Khmer native speaker),
    \item Javanese (Javanese--Indonesian bilingual, native Javanese speaker).
\end{itemize}

\paragraph{Task.} For each of 160~(source, instruction, candidate~translation) tuples per language, the annotator (a)~marked error spans (Minor / Major) on the candidate translation (b) assigned a 0--100 translation rating following the ESA protocol~\citep{kocmi-etal-2024-error}, and (c)~assigned a 0--100 adaptedness score reflecting how well the translation matches the instruction's purpose. The annotation interface (Figure~\ref{fig:annotation}) displays the source sentence, the instruction, and a single candidate translation; the source language (English) and target language were shown explicitly.

\paragraph{Training and rubric.}
Each annotator received a video walkthrough covering: the ESA error-span taxonomy and rating scale, the adaptedness rubric (scale guide reproduced in Figure~\ref{fig:annotation}), and use of the Label Studio interface. Annotators could ask follow-up questions before and during the task. No quality filter was applied beyond the initial walkthrough; agreement metrics in Table~\ref{tab:human-llm-agreement} thus reflect single-annotator judgements.

\paragraph{Compensation.}
Approximately US\$20/hour. This rate is above the national minimum wage in the relevant jurisdictions and above prevailing rates for similar annotation work in the annotators' countries of residence.

\paragraph{Consent.}
Annotators were informed of the purpose of the task (evaluation of machine-translation outputs for a research publication), the format of the data, and the use of their judgements (aggregated, non-identifying). Consent was obtained prior to annotation.

\end{document}